\documentclass[lettersize,journal]{IEEEtran}
\usepackage{amsmath,amsfonts}
\usepackage{algorithmic}
\usepackage{algorithm}
\usepackage{array}
\usepackage{textcomp}
\usepackage{stfloats}
\usepackage{url}
\usepackage{verbatim}
\usepackage{graphicx}
\usepackage{cite}
\usepackage{booktabs}
\usepackage{multirow}

\usepackage{listings}
\usepackage{colortbl}
\usepackage{subcaption}

\definecolor{codegreen}{rgb}{0,0.6,0}
\definecolor{codegray}{rgb}{0.5,0.5,0.5}
\definecolor{codepurple}{rgb}{0.58,0,0.82}
\definecolor{backcolour}{rgb}{0.95,0.95,0.92}
\definecolor{cvprblue}{rgb}{0.21,0.49,0.74}
\usepackage[pagebackref,breaklinks,colorlinks,citecolor=cvprblue]{hyperref}
\lstdefinestyle{pytorchstyle}{
    backgroundcolor=\color{white},
    commentstyle=\color{codegreen},
    keywordstyle=\color{magenta},
    numberstyle=\tiny\color{codegray},
    stringstyle=\color{codepurple},
    basicstyle=\ttfamily\small,
    breakatwhitespace=false,
    breaklines=true,
    captionpos=b,
    keepspaces=true,
    numbers=left,
    numbersep=5pt,
    showspaces=false,
    showstringspaces=false,
    showtabs=false,
    tabsize=2
}

\begin{document}

\title{DPStyler: Dynamic PromptStyler for Source-Free Domain Generalization}

\author{Yunlong Tang, Yuxuan Wan, Lei Qi, Xin Geng, Senior Member, IEEE
\thanks{The work is supported by NSFC Program (Grants No. 62206052, 62125602, 62076063), Jiangsu Natural Science Foundation Project (Grant No. BK20210224), China Postdoctoral Science Foundation (No. 2024M750424) and the Xplorer Prize.

Yunlong Tang (email: yltang1103@gmail.com), Yuxuan Wan (email: yaser@seu.edu.cn) and Xin Geng (email: xgeng@seu.edu.cn) are with the Key Laboratory of New Generation Artificial Intelligence Technology and Its Interdisciplinary Applications, Ministry of Education (Southeast University), China, 211189.

Lei Qi (e-mail: qilei@seu.edu.cn) is with the Key Laboratory of New Generation Artificial Intelligence Technology and Its Interdisciplinary Applications, Ministry of Education (Southeast University),  and the National Center of Technology Innovation for EDA, Nanjing, China, 211189.}
\thanks{Yunlong Tang and Yuxuan Wan are co-first authors. Corresponding authors: Lei Qi and Xin Geng.}}



\maketitle

\begin{abstract}
Source-Free Domain Generalization (SFDG) aims to develop a model that works for unseen target domains without relying on any source domain. Research in SFDG primarily bulids upon the existing knowledge of large-scale vision-language models and utilizes the pre-trained model's joint vision-language space to simulate style transfer across domains, thus eliminating the dependency on source domain images. However, how to efficiently simulate rich and diverse styles using text prompts, and how to extract domain-invariant information useful for classification from features that contain both semantic and style information after the encoder, are directions that merit improvement. In this paper, we introduce Dynamic PromptStyler (DPStyler), comprising Style Generation and Style Removal modules to address these issues. The Style Generation module refreshes all styles at every training epoch, while the Style Removal module eliminates variations in the encoder's output features caused by input styles. Moreover, since the Style Generation module, responsible for generating style word vectors using random sampling or style mixing, makes the model sensitive to input text prompts, we introduce a model ensemble method to mitigate this sensitivity. Extensive experiments demonstrate that our framework outperforms state-of-the-art methods on benchmark datasets. Code is available \href{https://github.com/TYLfromSEU/DPStyler}{here}.

\end{abstract}

\begin{IEEEkeywords}
Computer Vision, Domain Shift, Source-Free Domain Generalization.
\end{IEEEkeywords}

\section{Introduction}
\IEEEPARstart{D}{eep} neural networks achieve optimal performance under the fundamental assumption of independent and identically distributed training and testing data. However, when the target domain exhibits a different distribution from the source domain \cite{sariyildiz2021concept,wang2022revisiting,bengio2019meta,recht2019imagenet,hendrycks2018benchmarking,ye2021alleviating}, the effectiveness of these models significantly decreases. To tackle this issue, domain adaptation (DA) methods \cite{ben2006analysis,hoffman2018cycada,lee2022fifo,lee2022surgical,saito2019semi,sun2016return,tzeng2017adversarial,zhao2019learning,wang2022cross,lu2021discriminative} have been extensively studied. These methods assume access to the target domain and aim to adapt the model to that specific domain. Nonetheless, in many real-world scenarios, the target domain is inaccessible, leading to the domain generalization (DG) paradigm \cite{huang2021fsdr,yue2019domain,seo2020learning,kang2022style,yang2022domain,yang2022cycle,yang2022manydg,niu2023knowledge,jin2021style,fang2023three}. Domain generalization assumes access to one or more source domains but excludes access to the target domain. 
The primary goal of domain generalization is to improve the model's generalization ability to any unseen domain.
However, models trained on different source domains often exhibit varying degrees of generalization, making it challenging to determine which domains are beneficial for enhancing the model's generalization ability.  Moreover, the process of gathering and annotating extensive multi-source domain data for training purposes can be financially burdensome and, at times, impractical.
\begin{figure}
    \centering
    \includegraphics[width=1\columnwidth]{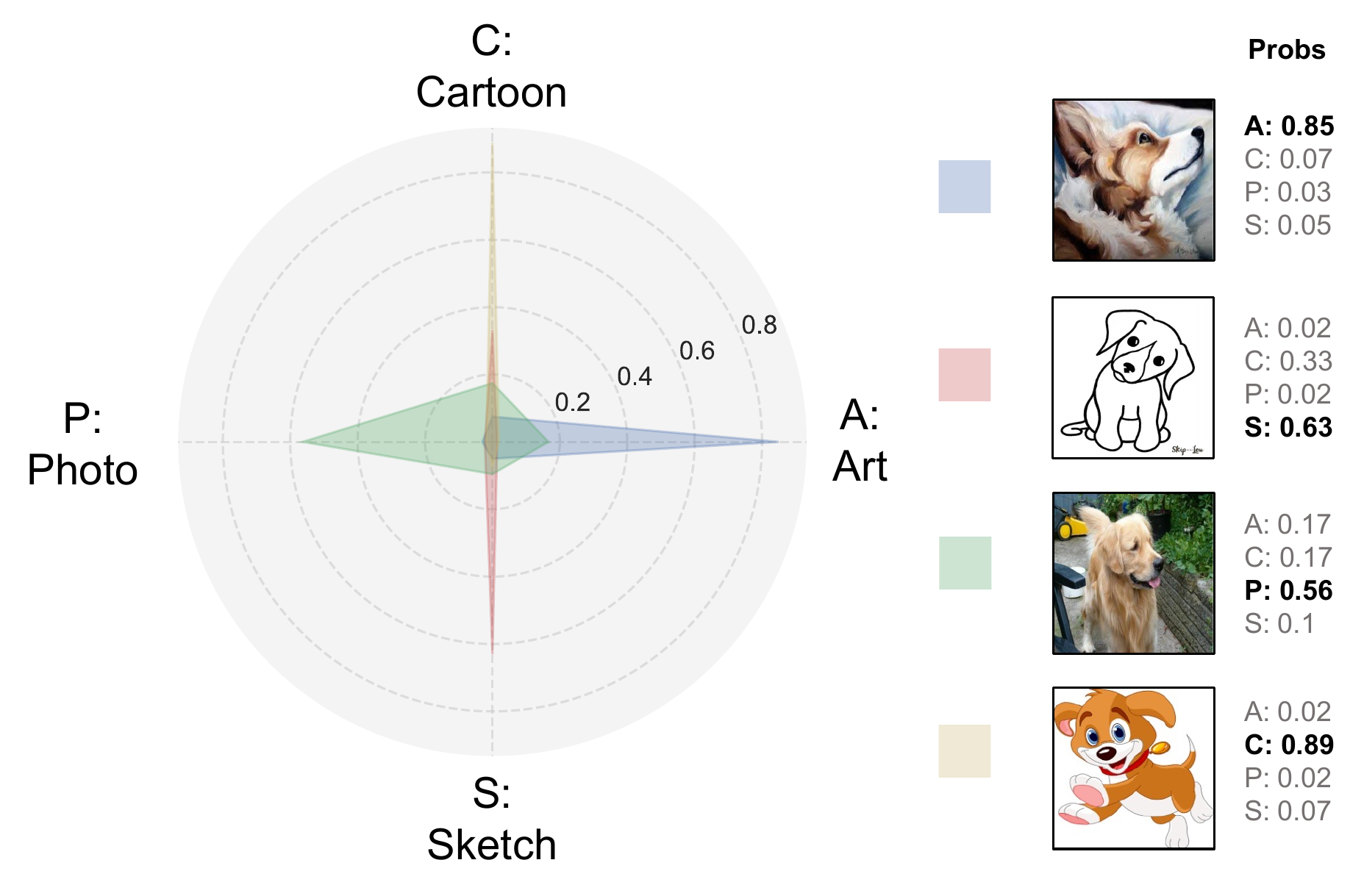}
    \caption{
    Domain classification via zero-shot using CLIP. 
    The four vertices of a color region represent the probabilities of the corresponding image's association with four domains, obtained through CLIP's zero-shot capabilities.  \textit{Image features stem from CLIP's image encoder}, while  text features are obtained from the descriptive text (\textit{eg}, ``a picture with a $\boldsymbol{\mathit{S}}$-like style," where $\boldsymbol{\mathit{S}}$ represents the style, such as ``cartoon") processed through the text encoder, representing the domain information. 
    Final probabilities are computed based on the similarity between image and text features. It can be found that the image features are domain-specific.
    }
    \label{figure1}
    \vspace{-0.5cm}
\end{figure}

Recently, Several methods have already tackled source-free domain generalization (SFDG) without using any images \cite{cho2023promptstyler,frikha2023towards,niu2022domain}. This task operates under the assumption that direct access to any actual data from the source or target domains is unavailable, solely relying on the definition of the target task, including class names.
Cho \textit{et al.} \cite{cho2023promptstyler} propose a prompt-driven style generation method called PromptStyler, which leverages the vast knowledge present in large-scale vision-language models (\textit{eg}, CLIP \cite{radford2021learning}), and the joint vision-language space to solve the problem. The approach involves training a classifier using text features while conducting inference with the classifier using image features. Before training this classifier, diverse styles are designed and embedded into text prompts, thereby enhancing the diversity of training samples.
However, there exist two issues: 1) PromptStyler first learns to obtain a certain number of styles and then keeps the styles fixed in subsequent classifier training. This restricts the model to only having access to a limited set of styles. 2) 
PromptStyler freezes the encoder and feeds the encoder's output directly to the classifier, making it difficult for the model to learn domain-invariant features.

\begin{figure}
    \centering
    \includegraphics[width=1\columnwidth]{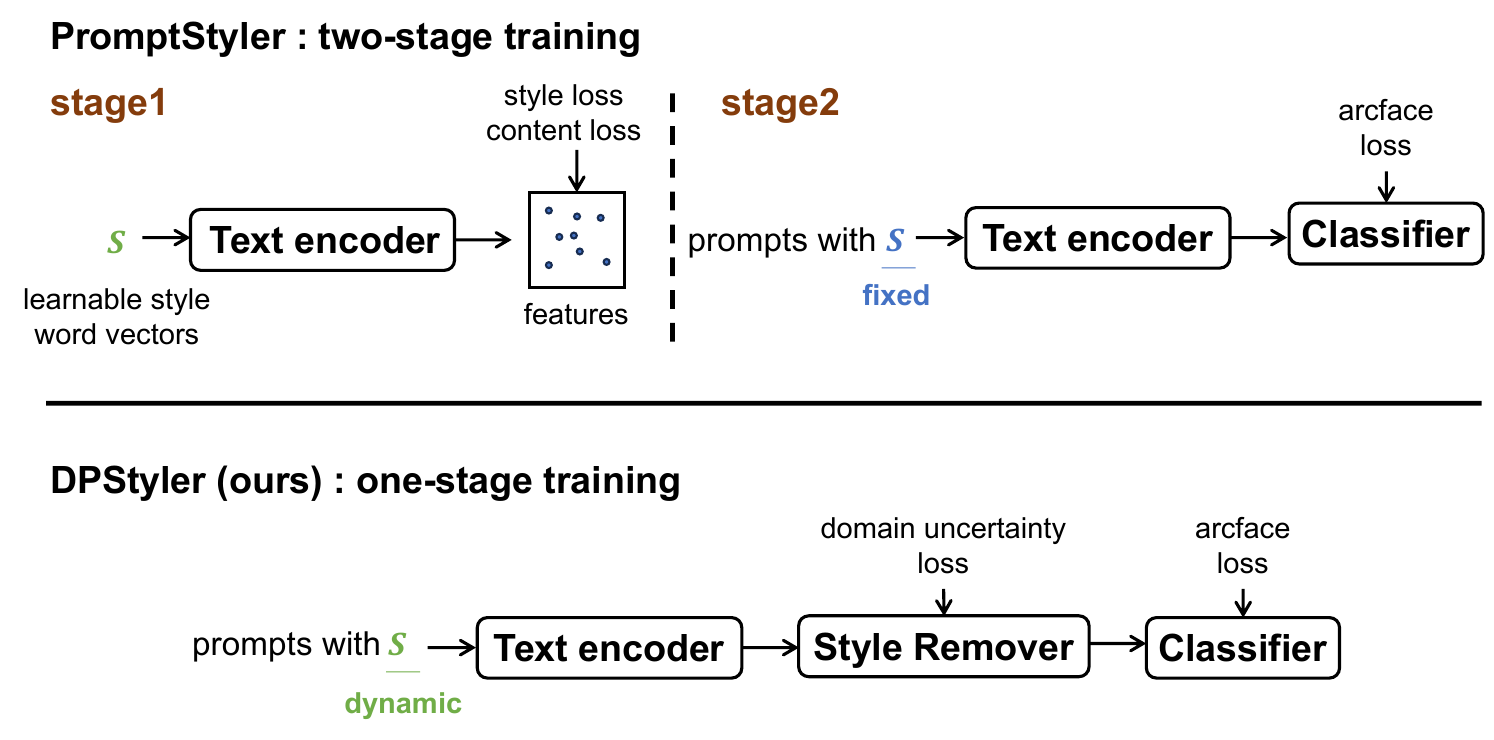}  
    \caption{
    The training strategies of PromptStyler~\cite{cho2023promptstyler} and DPStyler (Ours). PromptStyler requires two-stage training and fixes the styles in the second stage. Instead, ours requires only one-stage training and dynamically updates the styles during training.
    }
    \label{figure2}
    \vspace{-0.5cm}
\end{figure}

In this paper, we introduce \textbf{Dynamic PromptStyler (DPStyler)}, which is grounded in the task definition of source-free domain generalization. Leveraging the joint vision-language space of large-scale pre-trained multimodal models, our approach addresses domain shift problems through a one-stage training strategy, which is depicted in the Figure \ref{figure2}. 
We design our framework from the following perspectives:
\textit{1) How to achieve richer and more flexible styles?} Since fixed style word vectors cannot represent the wide variety of styles in the real world, we suggest dynamically refreshing styles by regenerating all style word vectors between epochs.  Two style generation methods are designed: Random and StyleMix. At each training epoch, we randomly select one method for refreshing styles to ensure a rich and diverse set of styles, resulting in \textbf{Style Generation Module}.
\textit{2) How to capture domain-invariant features?} 
Experimental results reveal that the output features from the image encoder still contain noticeable style information, negatively impacting classification performance. As depicted in the Figure \ref{figure1}, it is evident that the image output features are domain-specific. Thus, we introduce a \textbf{Style Removal Module} following the encoder. The aim is to reduce specific style information in the encoder output and learn domain-invariant features of objects. To guide the training of the style remover, we propose a domain uncertainty loss to ensure that the probability of the output feature having any style is similar. 


Besides, we observe that introducing more randomness into training process by dynamically refreshing styles makes the model more sensitive to changes in the input text prompts. To mitigate the impact of different text prompts on results, we employ \textbf{Model Ensemble} to improve the stability of the model. This involves training multiple models for distinct prompts, all of which contribute to the inference process. We demonstrate the effectiveness of our framework through experiments on four domain generalization benchmarks. Our framework consistently achieves state-of-the-art performance. Particularly, our method requires only one-stage training, different from PromptStyler, as shown in Figure \ref{figure2}.
Our contributions are summarized as follows:
\begin{itemize}
  \item  We propose an innovative and adaptable approach for SFDG that enhances the flexibility of the model by dynamically updating the styles during training.
  \item We propose a specialized style remover with a domain uncertainty loss incorporated after the encoder, encouraging the model's focus on domain-invariant features.
  \item  We employ a model ensemble technique to significantly enhance the stability of the model, effectively addressing the issue of sensitivity to input text prompts.
\end{itemize}

\section{Related Work}
\label{sec:related}
\subsection{Domain Generalization}
Improving and enhancing a model's generalization performance to unseen domains is a critical factor in the practical deployment of deep learning networks. This is because the out-of-distribution between different domains can significantly degrade a model's performance on the target domains. To address this issue, Domain Generalization (DG) has been investigated \cite{wang2022generalizing,carlucci2019domain,cha2022domain,gulrajani2020search,kim2021selfreg,zhou2020domain,zhou2020deep}. In contrast to Domain Adaptation (DA), DG assumes that the target domain is entirely inaccessible, and the model can only be trained using data from the source domains. Generally, most DG methods can be categorized into two groups: multi-source DG \cite{nam2021reducing,kang2022style,volpi2018generalizing,zhou2020learning,li2018domain,shao2019multi,yao2022pcl,li2019episodic} and single-source DG \cite{fan2021adversarially,li2021progressive,qiao2020learning,wang2021learning}. Multi-source DG methods typically aim to learn domain-invariant features across multiple distinct source domains to enhance the model's generalization performance. On the other hand, single-source DG, due to having only one source domain, often focuses on data augmentation based on this single domain to generate multiple synthetic domains, subsequently learns domain-invariant features from these synthesized domains.

\begin{figure*}[ht!]
    \centering
    \includegraphics[width=1\linewidth]{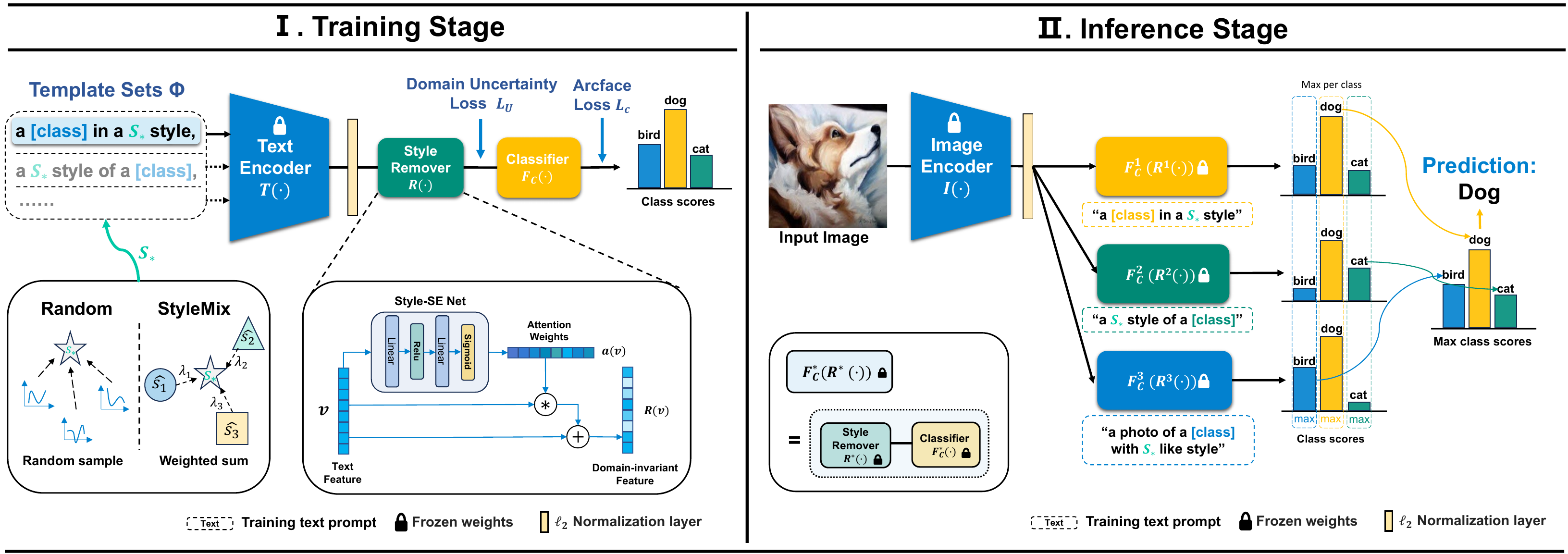}  
    \caption{The training process and inference process of DPStyler. A Style Generation Module is used to dynamically refresh styles during training with two style-refresh methods. A style remover with domain uncertainty loss is used to remove specific domain information and learn domain-invariant features. Model ensemble is used for the inference process. For the class scores generated by the models corresponding to different templates, we select the class corresponding to the maximum value from them as the prediction.}
    \label{figure3}
    \vspace{-0.4cm}
\end{figure*}
\vspace{-0.1cm}
\subsection{Vision-Language Models}
Several works have explored the joint learning of image and text representations \cite{radford2021learning,li2022blip,li2023blip,jia2021scaling,yang2022vision,zheng2022faster}. These works utilize image-text pairs for training visual-semantic embeddings, which find application not only in image classification, captioning, and retrieval but also in zero-shot prediction for unseen labels. For instance, CLIP \cite{radford2021learning} presents a scalable contrastive pre-training approach for learning joint text and image features. Leveraging a vast corpus of $400$ million image-text pairs, CLIP achieves superior zero-shot learning capabilities in classification tasks. Recent studies have demonstrated that leveraging style-specific prompts, such as `art,' `sketch,' or `photo' within their joint vision-language spaces allows for the manipulation of visual features \cite{dunlap2022using,gal2022stylegan,patashnik2021styleclip,kwon2022clipstyler}. Specially, Textual Inversion \cite{gal2022image} shows that a learnable style word vector for a pseudo-word $\boldsymbol{S}_*$ could capture a shared style of images using CLIP with a prompt (\textit{eg}, ``a painting in the style of $\boldsymbol{S}_*$"). 
\vspace{-0.1cm}
\subsection{Source-free Domain Generalization}
In recent years, a new task called source-free domain generalization without using any images has been proposed \cite{cho2023promptstyler,frikha2023towards,niu2022domain}. In this task, we are not allowed to access any data from the source domain or the target domain, and only the target task definition, such as class names, is provided. Therefore, source-free domain generalization is much difficult but also more closely related to real-world applications. To address this problem, Cho \textit{et al.} \cite{cho2023promptstyler} propose a prompt-driven style generation method called PromptStyler. This method synthesizes diverse styles by using learnable word vectors to simulate distribution shifts in a hyperspherical joint vision-language space.
In particular, the method trains to produce a variety of style features (from ``a $\boldsymbol{S}_{*}$ style of a") by leveraging trainable vectors representing pseudo-words $\boldsymbol{S}_{*}$. To safeguard against potential distortion of content information, PromptStyler enforces the proximity of style-content features (from ``a $\boldsymbol{S}_{*}$ style of a [class]") to their associated content features (from ``[class]") within the joint vision-language space.
Then, a classifier is trained using text features and conducts inference using image features. Here, we improve upon PromptStyler by introducing novel style word vector generation strategies. This transforms the method from a two-stage approach to a one-stage approach, no longer need to train style word vectors individually. Meanwhile,  a style remover is proposed to capture domain-invariant features.

\section{Method}
The overall framework of the proposed method is shown in Figure \ref{figure3}, and the pseudo-code is described in Algorithm \ref{alg:1}. 
Our approach leverages a large pre-trained model's  joint vision-language space (\textit{eg}, CLIP \cite{radford2021learning} latent space), trains with the text encoder $T(\cdot)$ and inferences with the image encoder $I(\cdot)$ . During training, we create diverse prompts as inputs to the text encoder and employ a style removal module $R(\cdot)$ to learn domain-invariant features for classification. During inference, we simply transplant the trained style remover and classifier to the image encoder for classification tasks. Moreover, model ensemble is employed to enhance the stability of the model. Note that we exploit CLIP as our large-scale vision-language model. All parameters in the CLIP model are frozen in our entire framework.
\vspace{-0.15cm}
\subsection{Style Generation Module}
\vspace{-0.06cm}
Since no source domain data can be accessed, we need to create a set of text prompts $\{\mathcal{P}\}$ in advance to approximate the representations of the test images we will encounter.
Suppose that the initial text template $\mathcal{P}_{m}^{\mathrm{content}} \circ \mathcal{P}_{i}^{\mathrm{style}}$ to be ``a [class]\textsubscript{m} in a $\boldsymbol{S}_i$ style'', where ``[class]\textsubscript{$m$}" denotes the $m$-th class name, and ``$\boldsymbol{S}_i$'' represents the placeholder for the $i$-th style word vector. Prior to encoding, an input text prompt undergoes tokenization to generate multiple tokens, which are subsequently substituted with their respective word vectors through a word lookup mechanism.  The pseudo-word $\boldsymbol{S}_i$ in a prompt is replaced by a style word vector $\mathbf{s}_i \in \mathbb{R}^D$ during the word lookup process. Assuming there are $M$ classes defined by the task , and we want to generate $K$ different styles, the number of training samples equals $M\times K$.

In PromptStyler, learnable style word vectors $\{ \mathbf{s}_i \}^K_{i=1}$ are initially sampled from a normal distribution, and then style diversity loss and content consistency loss are applied to achieve diverse styles while maintaining semantic information.
Instead, we propose a novel style generation strategy in which all styles are refreshed at the beginning of each training epoch. This integrates the style generation process with the model training process and allows the model to continuously encounter new style information. Note that all these style word vectors $\{ \mathbf{s}_i \}^K_{i=1}$ do not need to be obtained by training. We have designed two style refresh methods: Random Generation, StyleMix Generation.

\noindent\textbf{Random Generation.} In our initial setup, we assume five different distributions: normal, xavier uniform, xavier normal, kaiming normal, kaiming uniform. For each style update, we randomly select one distribution from the five and sample a new style word vector from it as an updated style. Repeating this process $K$ times refreshes all styles.

\noindent\textbf{StyleMix Generation.} Randomly generated styles may not necessarily contain meaningful semantic information. For instance, a randomly generated word vector might have a semantic similarity closer to the word ``this". To address this issue, we propose a style-mix generation method. Initially, a set of adjectives are predefined, such as ``white", ``cartoon", and others. Then the corresponding word vectors $\{\hat{\mathbf{s}} \in \mathbb{R}^D\}$ for these vocabulary words  are obtained by token embedding. The new style word vector is derived through a weighted sum of the predefined word vectors:
\begin{equation}
\mathbf{s}_i=\sum \lambda_j \hat{\mathbf{s}}_j\,,
\end{equation}
where $\hat{\mathbf{s}}_j$ represents the $j$-th predefined word vector, and $\lambda_j$ represents the corresponding weight coefficient which is sampled from a beta distribution and satisfies $\sum \lambda_j = 1$.

At each epoch's style refresh, we randomly select one of the methods with a $50\%$ probability for regenerating styles.
\begin{figure}
    \centering
    \includegraphics[width=1\linewidth]{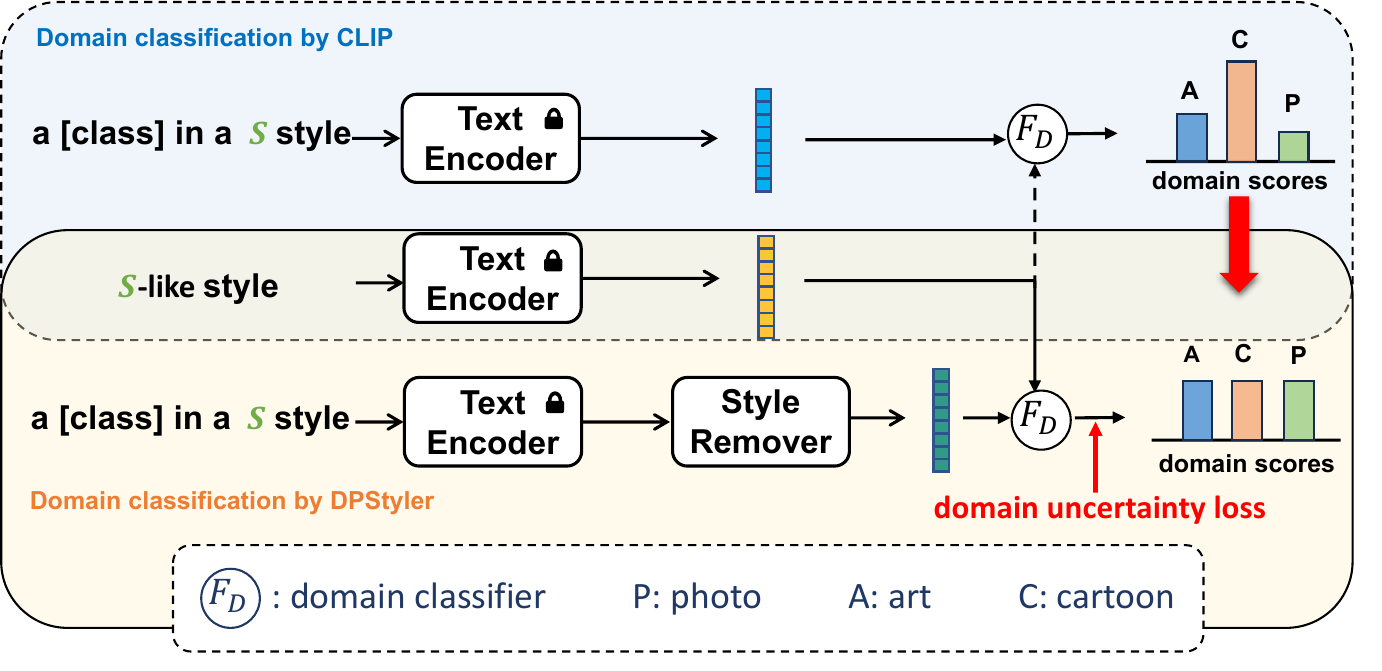}  
    \caption{Illustration of domain uncertainty loss. It is used to constrain style remover to remove domain-specific information.}
    \label{figure_LU}
    \vspace{-0.5cm}
\end{figure}

\subsection{Style Removal Module}
After feeding the text prompt $\{\mathcal{P}\}$ into the text encoder $T(\cdot)$, object features with diverse style information $\{T(\mathcal{P}_{m}^{\mathrm{content}} \circ \mathcal{P}_{i}^{\mathrm{style}}) \in \mathbb{R}^C\}_{m=1,i=1}^{M,K}$ will be obtained. Subsequently, a style remover $R(\cdot)$ is employed to remove the specific style information from the features and obtain domain invariant features for better classification. In this paper we employ SE-like model Style-SE Net as the style remover.  This module utilizes two fully connected layers to form a bottleneck structure and is implemented in residual form. Specifically, for the input feature $T(\mathcal{P}_{m}^{\mathrm{content}} \circ \mathcal{P}_{i}^{\mathrm{style}}) \in \mathbb{R}^C$, the number of channels $C$ is first reduced to $C/r$ through a fully connected layer, and then restored to $C$ after relu activation by another fully connected layer, where $r$ is feature compression ratio. Subsequently, a sigmoid operation is applied to obtain the weights for each channel, which are multiplied with the corresponding channel values. In summary, the style remover can be written as:
\begin{equation}
a(\mathbf{v})=\operatorname{Sigmoid}(\operatorname{ReLU}(\mathbf{v}^T \mathbf{W}_1) \mathbf{W}_2)\,,
\end{equation}
\begin{equation}
R(\mathbf{v})=a(\mathbf{v})*\mathbf{v}+\mathbf{v}\,,
\end{equation}
where $\mathbf{v}$ denotes the input to this module and $\mathbf{W}_1,\mathbf{W}_2$ denote the weights of the two fully connected layers. The features output by the image encoder contain both semantic and style information, and our goal is to decouple these two types of information, retaining the semantic information that is useful for classification. So the objective of Style-SE Net is to highlight semantic information and reduce the influence of style information by assigning different weights to different channels of the features, which means channels that represent style information will be given smaller weights.

\begin{algorithm}[t] 

\caption{Pseudocode of DPStyler in a PyTorch style}
\label{alg:1}
\begin{lstlisting}[language=Python, basicstyle=\scriptsize\ttfamily]
"""
T: Text Encoder,R: Style Removal Module
F_C: class Classifier, F_D: Domain Classifier
style: style embedding list, dim is (K, 1, C)
L_u: domain uncertainty loss function, L_c: classification loss function
x: text prompt (e.g., "a S* style of a [class]") embedding list, dim is (K*M, L, C), where L is the length of the text prompt.
"""
for epoch in range(epochs):
    # Apply random or mix style generation strategy to update style embedding
    random_choice = random.randint(0, 1)
    if random_choice == 0:
        style = random_generator() # Random style generation strategy
    else:
        style = mix_generator() # Mix style generation strategy
    for (x, y) in train_loader:
        # update content_style prompt embedding
        x[:, style_idx: style_idx+1, :] = style
        # Output of the text encoder
        t_out = T(x) 
        # Remove the style by passing through the Style Removal Module
        t_out_r = R(t_out)
        # Predict class score
        cls_score = F_C(t_out_r)
        # Predict domain score
        domain_score = F_D(t_out_r)
        # Compute losses
        loss = L_u(domain_score) + L_c(cls_score, y)
        loss.backward()
\end{lstlisting}
\end{algorithm}

To effectively remove specific style information with this module, we utilize a domain classifier $F_D(\cdot)$ and a domain uncertainty loss $\mathcal L_U$ to guide the training of Style-SE Net. Firstly, a style prompt template $\mathcal{P}_{i}^{\mathrm{style}}$ (\textit{eg},  ``$\boldsymbol{S}_i$-like style"), is initialized. Then we create various style prompts $\{\mathcal{P}_{i}^{\mathrm{style}}\}_{i=1}^K$ by replacing $\boldsymbol{S}_i$ with the $K$ style word vectors obtained by the style generation module. These style prompts are encoded by text encoder $T(\cdot)$ to get the text features $\{T(\mathcal{P}_{i}^{\mathrm{style}})\in \mathbb{R}^C\}_{i=1}^K$ corresponding to the $K$ styles. Similar to CLIP's zero-shot prediction, the input of $F_D(\cdot)$ consists of features $R(T(\mathcal{P}_{m}^{\mathrm{content}} \circ \mathcal{P}_{i}^{\mathrm{style}})) \in \mathbb{R}^C$ obtained by style remover and the style text features $\{T(\mathcal{P}_{i}^{\mathrm{style}})\in \mathbb{R}^C\}_{i=1}^K$, the output is probabilities computed by consine similarity. Thus, the probability that a sample has a certain style $\boldsymbol{S}_j$ can be written as: 
\begin{equation}
z_{m i j}=\frac{R(T(\mathcal{P}_m^{\text {content }} \circ \mathcal{P}_i^{\text {style }}))}{\|R(T(\mathcal{P}_m^{\text {content }} \circ \mathcal{P}_i^{\text {style }}))\|_2} \cdot \frac{T(\mathcal{P}_j^{\text {style }})}{\|T(\mathcal{P}_j^{\text {style }})\|_2} \,,
\end{equation}

\begin{equation}
p_{m i j}=\frac{e^{z_{m i j }}} {\sum_{k=1}^{K} e^{z_{m i k}}}\,.
\end{equation}
Then, for a sample $\mathcal{P}_m^{\text {content }} \circ \mathcal{P}_i^{\text {style }}$, 
the output of domain classifier $F_D(\cdot)$ can be represented as $\{p_{m i j}\}_{j=1}^K$ .

As mentioned above, the features output by the encoder still contain obvious style information. We hope that after style removal, the similarity between the output feature and various style features is similar, \textit{ie}, it is difficult to discriminate which domain the sample belongs to, as depicted in Figure \ref{figure_LU}. Therefore, based on $F_D(\cdot)$, we use the entropy loss to realize the domain uncertainty loss $\mathcal L_U$:
\begin{equation}
\mathcal L_U(\mathcal{P}_{m}^{\mathrm{content}} \circ \mathcal{P}_{i}^{\mathrm{style}})=\sum_{j=1}^{K} p_{m i j} \log p_{m i j}\,,
\end{equation}
where $p_{m i j}$ is the probability that sample $\mathcal{P}_{m}^{\mathrm{content}} \circ \mathcal{P}_{i}^{\mathrm{style}}$ has style $\boldsymbol{S}_i$ and $K$ is the number of styles created. The domain uncertainty loss aims to diminish the specific expression of style information in the features, thereby suppressing the channels associated with style information and implicitly highlighting the domain-invariant information.
\begin{table*}
\centering
\caption{Comparison with state-of-the-art domain generalization methods, ZS-CLIP (C) refers to zero-shot CLIP using `[class]' as its text prompt, and ZS-CLIP (PC) indicates zero-shot CLIP using `a photo of a [class]' as its text prompt. 
\textbf{Bold} indicates the best result. `*' denotes our reproduced results for PromptStyler, and `\dag' denotes that the paper corresponding to the method does not give the standard errors. }
 \resizebox{\linewidth}{!}{
\begin{tabular}{l|c|ccccc} 
\toprule
\multicolumn{1}{l|}{\multirow{2}{*}{Method}} &  \multirow{2}{*}{Venue}  & \multicolumn{5}{c}{Accuracy (\%) $\uparrow$}                                                                      \\ 
\cline{3-7}
\multicolumn{1}{c|}{}                        &                        & \quad\makebox[0.1\textwidth][c]{PACS}              & \makebox[0.1\textwidth][c]{VLCS}              &\makebox[0.1\textwidth][c]{OfficeHome}      & \makebox[0.1\textwidth][c]{DomainNet}          &  \makebox[0.1\textwidth][c]{Avg.}           \\ 
\hline
\multicolumn{7}{c}{ResNet-50 with pre-trained weights on ImageNet}                                                                                                             \\ 
\hline
DANN \cite{ganin2016domain} & \makebox[0.14\textwidth][c]{JMLR'2016} &  \quad\makebox[0.1\textwidth][c]{85.2 ± 0.9}        & \makebox[0.1\textwidth][c]{77.1 ± 0.5}       & \makebox[0.1\textwidth][c]{65.5 ± 0.9}         & \makebox[0.1\textwidth][c]{38.9 ± 0.5}      & \makebox[0.1\textwidth][c]{66.7}       \\ 
RSC \cite{huang2020self}                                          & ECCV'2020            & \quad \makebox[0.1\textwidth][c]{85.2 ± 0.9}   & \makebox[0.1\textwidth][c]{77.1 ± 0.5}        & \makebox[0.1\textwidth][c]{65.5 ± 0.9}         & \makebox[0.1\textwidth][c]{38.9 ± 0.5}       & \makebox[0.1\textwidth][c]{66.7}      \\
LP-FT   &  ICLR'2021  & \quad84.6 ± 0.8           &  76.7 ± 1.5  &  65.0 ± 0.2  &  43.0 ± 0.1  &  67.3   \\ 
MLDG \cite{li2018learning}                                         &  AAAI'2018              & \quad\makebox[0.1\textwidth][c]{84.9 ± 1.0}         & \makebox[0.1\textwidth][c]{77.2 ± 0.4}        & \makebox[0.1\textwidth][c]{66.8 ± 0.6}         & \makebox[0.1\textwidth][c]{ 41.2 ± 0.1 }        & 67.5        \\
SagNet \cite{nam2021reducing}                                       &  CVPR'2021              &  \quad86.3 ± 0.2  &  77.8 ± 0.5           &  68.1 ± 0.1           &  40.3 ± 0.1           &  68.1            \\
SelfReg \cite{kim2021selfreg}                                      &  ICCV'2021              &  \quad85.6 ± 0.4           &  77.8 ± 0.9           &  67.9 ± 0.7           &  42.8 ± 0.0           &  68.5            \\
GVRT \cite{min2022grounding}                                        &  ECCV'2022             &  \quad85.1 ± 0.3           &  79.0 ± 0.2  &  70.1 ± 0.1           &  44.1 ± 0.1           &  69.6            \\ 
MIRO \cite{cha2022domain}                                         & ECCV'2022              &  \quad85.4 ± 0.4           &  79.0 ± 0.0  & 70.5 ± 0.4  &  44.3 ± 0.2  &  69.8  \\ 
Text Regularization$^{\dag}$ \cite{zhang2023learning}                                         & PRCV'2023              &  \quad87.2 ± 0.0           &  \textbf{80.3 ± 0.0}  &  70.4 ± 0.4  &  44.0 ± 0.0  &  70.5   \\ 
Model Ratatouille$^{\dag}$ \cite{rame2023model}  & ICML'2023   & \quad\textbf{89.8 ± 0.0}           &  78.3 ± 0.0  &  \textbf{73.5 ± 0.0}  &  \textbf{47.7 ± 0.0}  &  \textbf{72.3}   \\

\hline
\multicolumn{7}{c}{ResNet-50 with pre-trained weights from CLIP}                                                                                                               \\ 
\hline
ZS-CLIP (C) \cite{radford2021learning}                                 & ICML'2021              &  \quad90.6 ± 0.0           &  79.4 ± 0.0           &  67.4 ± 0.0           &  45.9 ± 0.0           &  70.8            \\
CAD  \cite{ruan2021optimal}                                         &  ICLR'2022           &  \quad90.0 ± 0.6           &  81.2 ± 0.6           &  70.5 ± 0.3           &  45.5 ± 2.1           &  71.8            \\
ZS-CLIP (PC) \cite{radford2021learning}                                 &  ICML'2021              &  \quad90.7 ± 0.0           &  82.0 ± 0.0           &  71.1 ± 0.0           &  46.1 ± 0.0           &  72.5            \\ 
Text Regularization$^{\dag}$ \cite{zhang2023learning}                                         & PRCV'2023              &  \quad91.3 ± 0.0           &  82.8 ± 0.0  &  71.6 ± 0.4  &  44.6 ± 0.0  &  72.6   \\ 
PromptStyler* \cite{cho2023promptstyler}                              &  ICCV'2023              &  \quad93.1 ± 0.4           &  82.2 ± 0.2           &  71.0 ± 0.1           &  46.9 ± 0.2           &  73.3            \\ \hline
\rowcolor[rgb]{0.929,0.929,0.929} DPStyler   &  Ours                  &  \quad\textbf{93.6 ± 0.2}   &  \textbf{83.5 ± 0.2}  &  \textbf{72.5 ± 0.2}  &  \textbf{48.0 ± 0.1}  & \makebox[0.1\textwidth][c]{$\textbf{74.4}$}   \\ 
\hline
\multicolumn{7}{c}{ViT-B / 16 with pre-trained weights from CLIP}                                                                                                              \\ 
\hline
ZS-CLIP (C) \cite{radford2021learning}                                  & ICML'2021            & \quad 95.6 ± 0.0           &  76.2 ± 0.0           &  79.6 ± 0.0           &  57.4 ± 0.0           &  77.2            \\
MIRO$^{\dag}$ \cite{cha2022domain}                                          &  ECCV'2022            &  \quad95.6 ± 0.0         &  82.2 ± 0.0           &  82.5 ± 0.0           &  54.0 ± 0.0           &  78.6            \\
ZS-CLIP (PC) \cite{radford2021learning}                                 &  ICML'2021              &  \quad96.0 ± 0.0           &  83.0 ± 0.0           &  81.8 ± 0.0           &  57.2 ± 0.0           &  79.5            \\ 
PromptStyler* \cite{cho2023promptstyler}                                &  ICCV'2023              & \quad 96.8 ± 0.2           &  83.7 ± 0.5  &  81.8 ± 0.4           &  56.7 ± 0.3           &  79.8            \\ 
Text Regularization$^{\dag}$ \cite{zhang2023learning}                                         & PRCV'2023              &  \quad95.9 ± 0.0           &  83.0 ± 0.0  &  82.3 ± 0.0  &  57.9 ± 0.0  &  79.8   \\ \hline
\rowcolor[rgb]{0.929,0.929,0.929} DPStyler   & Ours                  &  \quad\textbf{97.1 ± 0.1}  &  \textbf{84.0 ± 0.4}  &  \textbf{82.8 ± 0.1}  &  \textbf{58.9 ± 0.1}  &  \makebox[0.1\textwidth][c]{$\textbf{80.7}$}   \\ 
\hline
\multicolumn{7}{c}{ViT-L / 14 with pre-trained weights from CLIP}                                                                                                              \\ 
\hline
ZS-CLIP (C) \cite{radford2021learning}                                  & ICML'2021              & \quad 97.6 ± 0.0           &  77.5 ± 0.0           &  85.7 ± 0.0           &  63.1 ± 0.0           &  81.0            \\
ZS-CLIP (PC) \cite{radford2021learning}                                 &  ICML'2021             &  \quad98.3 ± 0.0           &  81.9 ± 0.0           &  86.6 ± 0.0           &  63.0 ± 0.0           &  82.5            \\
PromptStyler* \cite{cho2023promptstyler}                                &  ICCV'2023              & \quad \textbf{98.4 ± 0.1}  &  81.3 ± 0.3           &  86.4 ± 0.2           &  62.9 ± 0.2                    &  82.2               \\ \hline
\rowcolor[rgb]{0.929,0.929,0.929} DPStyler   & Ours                   &  \quad\textbf{98.4 ± 0.1}  &  \textbf{83.2 ± 0.1}  & \textbf{88.0 ± 0.3}  &  \textbf{64.7 ± 0.1}                    &  \makebox[0.1\textwidth][c]{$\textbf{83.6}$}              \\
\bottomrule
\end{tabular}
}

\label{tab:table1}
\vspace{-0.2cm}
\end{table*}
\subsection{Model Ensemble}
During our experiments we find that introducing more randomness into the training process makes the model more sensitive to changes in the input text prompts. Different initial text templates will make the model behave differently on each dataset, 
even if they are semantically similar (\textit{eg}, ``a [class] in a $\boldsymbol{S}_*$ style" and ``a $\boldsymbol{S}_*$ style of a [class]"). This lead us to think about how to reduce the impact of template selection on the model's effectiveness and harvest a stable and desirable classification result. We propose to integrate the advantages of each template in a model ensemble approach \cite{lin2022ensemble}. Specifically, \textbf{for each text template, we train a model and each model is involved in the inference process}, as depicted in the Figure \ref{figure3}. Assuming that $N$ models are trained, each model generates $M$ class scores during inference, yielding a total of $N\times M$ class scores. We select the class corresponding to the largest score value among these class scores as the predicted class for the input image.
\subsection{Model Training and Inference}
After obtaining the domain-invariant feature through a style remover, we use a linear classifier $F_C(\cdot)$ to obtain the class predictions. Similar to promptstyler, we utilize ArcFace \cite{deng2019arcface} Loss as our classification loss $\mathcal L_C$ to effectively leverage the hyperspherical joint vision-language space.
The goal of ArcFace loss is to ensure that features of the same class are closer together in the feature space, while keeping features of different classes dispersed. The total loss for the model training can be computed by:
\begin{equation}
\vspace{-0.05cm}
\mathcal L_{total}=\mathcal L_U+\mathcal L_C\,.
\vspace{-0.05cm}
\end{equation}
In the model inference process, utilizing the joint vision-language space, we simply transplant the trained style remover as well as classifier to the image encoder afterward. Specifically, for an input image $\mathbf{x}$, the image encoder extracts its image feature $I(\mathbf{x})\in \mathbb{R}^C$, which is mapped to the hyperspherical joint vision-language space by $\ell_2$ normalization. Its domain invariant feature $R(I(\mathbf{x}))\in \mathbb{R}^C$ is then obtained by the style remover and used for classification. The final prediction is generated by model ensemble.

\textbf{Remark:} Compared with PromptStyler, the advantages of our method include: \textit{1) Our method can further improve the performance with fewer training resources; 2) Our method simplifies the training pipline; 3) Our method does not need template selection owing to model ensemble. }

\begin{table*}
\centering
\caption{Comparison with SOTA domain generalization methods for per-domain top-1 accuracy on PACS. We repeat each experiment using three different seeds, and report average accuracies with standard errors. ZS-CLIP (C) refers to zero-shot CLIP using `[class]' as its text prompt, and ZS-CLIP (PC) indicates zero-shot CLIP using `a photo of a [class]' as its text prompt. 
\textbf{Bold} indicates the best result. `*' denotes our reproduced results for PromptStyler, and `\dag' denotes that the paper corresponding to the method does not give the standard errors.}
    \resizebox{\linewidth}{!}{
\begin{tabular}{l|c|ccccc} 
\toprule
\multicolumn{1}{l|}{\multirow{2}{*}{Method}} &  \multirow{2}{*}{Venue}  & \multicolumn{5}{c}{Accuracy (\%) $\uparrow$}                                                                      \\ 
\cline{3-7}
\multicolumn{1}{c|}{}                        &                        & \quad\makebox[0.1\textwidth][c]{Art Painting}              & \makebox[0.1\textwidth][c]{Cartoon}              &\makebox[0.1\textwidth][c]{Photo}      & \makebox[0.1\textwidth][c]{Sketch}          &  \makebox[0.1\textwidth][c]{Avg.}           \\ 
\hline
\multicolumn{7}{c}{ResNet-50 with pre-trained weights on ImageNet}                                                                                                             \\ 
\hline

GVRT \cite{min2022grounding}                                        &  ECCV'2022             &  \quad 87.9 ± 0.3           &  78.4 ± 1.0  &  98.2 ± 0.1           &  75.7 ± 0.4           &  85.1            \\ 
SAGM$^{\dag}$ \cite{wang2023sharpness}  &  CVPR'2023  &  \quad 85.3 ± 0.0           &  80.6 ± 0.0  &  95.9 ± 0.0           &  80.1 ± 0.0           &  85.5           \\ 
SelfReg \cite{kim2021selfreg}                                      &  \makebox[0.14\textwidth][c]{ICCV'2021}             &  \quad\makebox[0.1\textwidth][c]{87.9 ± 1.0}           &  \makebox[0.1\textwidth][c]{79.4 ± 1.4}           &  \makebox[0.1\textwidth][c]{96.8 ± 0.7}           &  \makebox[0.1\textwidth][c]{78.3 ± 1.2}           &  \makebox[0.1\textwidth][c]{85.6}            \\ 
M$^2$-CL$^{\dag}$ \cite{ballas2024multi}  &  TAI'2024  &  \quad 87.3 ± 0.0           &  81.8 ± 0.0  &  98.5 ± 0.0           &  76.3 ± 0.0           &  86.0           \\ 
JPDA$^{\dag}$ \cite{chen2023domain}  &  PR'2023  &  \quad \textbf{90.6 ± 0.0}           &  83.6 ± 0.0  &  \textbf{99.0 ± 0.0}           &  82.8 ± 0.0           &  89.0           \\ 
Model Ratatouille$^{\dag}$ \cite{rame2023model}  & ICML'2023   & \quad\textbf{90.6 ± 0.0}           &  \textbf{84.7 ± 0.0}  &  98.8 ± 0.0  &  \textbf{85.0 ± 0.0}  &  \textbf{89.8}   \\

\hline
\multicolumn{7}{c}{ResNet-50 with pre-trained weights from CLIP}                                                                                                               \\ 
\hline
ZS-CLIP (C) \cite{radford2021learning}                                 & ICML'2021              &  \quad 88.9 ± 0.0           &  94.4 ± 0.0           &  99.3 ± 0.0           &  79.8 ± 0.0           &  90.6            \\
ZS-CLIP (PC) \cite{radford2021learning}                                 &  ICML'2021              &  \quad 90.8 ± 0.0           &  93.3 ± 0.0           &  99.4 ± 0.0           &  79.3 ± 0.0           &  90.7           \\ 
PromptStyler* \cite{cho2023promptstyler}                              &  ICCV'2023              &  \quad 93.6 ± 0.3         & 95.2 ± 0.2           &  99.3 ± 0.0           &  84.2 ± 1.3           &  93.1            \\ \hline
\rowcolor[rgb]{0.929,0.929,0.929} DPStyler   &  Ours                  &  \quad\textbf{94.0 ± 0.2}   &  \textbf{95.3 ± 0.0}  &  \textbf{99.6 ± 0.0}  &  \textbf{85.5 ± 0.4}  & \makebox[0.1\textwidth][c]{$\textbf{93.6}$}   \\ 
\hline
\multicolumn{7}{c}{ViT-B / 16 with pre-trained weights from CLIP}                                                                                                              \\ 
\hline
ZS-CLIP (C) \cite{radford2021learning}                                  & ICML'2021            & \quad 96.4 ± 0.0           &  98.7 ± 0.0           &  \textbf{99.9 ± 0.0}           &  87.5 ± 0.0           &  95.6            \\
ZS-CLIP (PC) \cite{radford2021learning}                                 &  ICML'2021              &  \quad 97.3 ± 0.0           &  99.0 ± 0.0           &  \textbf{99.9 ± 0.0}          &  88.0 ± 0.0           &  96.1            \\
PromptStyler* \cite{cho2023promptstyler}                                &  ICCV'2023              & \quad 97.4 ± 0.2           &  \textbf{99.1 ± 0.1}  &  \textbf{99.9 ± 0.1}           &  90.7 ± 0.2           &  96.8            \\ \hline
\rowcolor[rgb]{0.929,0.929,0.929} DPStyler   & Ours                  &  \quad\textbf{97.7 ± 0.0}  &  98.9 ± 0.1  &  \textbf{99.9 ± 0.0}  &  \textbf{91.9 ± 0.5}  &  \makebox[0.1\textwidth][c]{$\textbf{97.1}$}   \\ 
\hline
\multicolumn{7}{c}{ViT-L / 14 with pre-trained weights from CLIP}                                                                                                              \\ 
\hline
ZS-CLIP (C) \cite{radford2021learning}                                  & ICML'2021              & \quad 97.2 ± 0.0           &  99.4 ± 0.0           &  99.9 ± 0.0           &  93.9 ± 0.0           &  97.6            \\
ZS-CLIP (PC) \cite{radford2021learning}                                 &  ICML'2021             &  \quad 98.6 ± 0.0           &  99.5 ± 0.0           &  99.9 ± 0.0           &  95.3 ± 0.0           &  98.3            \\
PromptStyler* \cite{cho2023promptstyler}                                &  ICCV'2023              & \quad \textbf{98.8 ± 0.3}  &  \textbf{99.8 ± 0.0}           &  \textbf{100.0 ± 0.0}           &  95.1 ± 0.5                    &  \textbf{98.4}               \\ \hline
\rowcolor[rgb]{0.929,0.929,0.929} DPStyler   & Ours                   &  \quad 98.5 ± 0.2  &  99.6 ± 0.1  & \textbf{100.0 ± 0.0}  &  \textbf{95.6 ± 0.2}                    &  \makebox[0.1\textwidth][c]{$\textbf{98.4}$}              \\
\bottomrule
\end{tabular}
}
\vspace{-0.2cm}

\label{tab:table_pacs}

\end{table*}
\section{Experiments}
\subsection{Evaluation Datasets}
To evaluate the effectiveness of our model, we conduct experiments on four domain generalization benchmarks: PACS  \cite{DBLP:conf/iccv/PACS} ($4$ domains, $7$ classes, and $9,991$ images), VLCS  \cite{DBLP:conf/iccv/VLCS} ($4$ domains, $5$ classes, and $10,729$ images), OfficeHome  \cite{DBLP:conf/cvpr/OfficeHome} ($4$ domains, $65$ classes, and $15,588$ images), and DomainNet  \cite{DBLP:conf/iccv/DomainNet} ($6$ domains, $345$ classes, and $586,575$ images). According to the task definition of source-free domain generalization, we do not use any source domain data for training, which is different from the \textit{leave-one-domain-out cross-validation} evaluation protocol \cite{DBLP:conf/iclr/protocol}.
\vspace{-0.1cm}
\subsection{Implementation Details}
\label{subsec:4.2}
DPStyler maintains consistent implementation and training with identical configurations across all evaluation datasets. The training process typically requires about $1$ minute when conducted on a single RTX3090 GPU, which is $\sim4.5\times$ faster compared to PromptStyler \cite{cho2023promptstyler}. Further detailed comparisons are available in Section \ref{sec:4.3}.

\textbf{Architecture.} We choose CLIP \cite{radford2021learning}  as our large-scale pretrained vision-language model, making use of the publicly available pre-trained model\footnote{\href{https://github.com/openai/CLIP}{https://github.com/openai/CLIP}}.
The text encoder $T(\cdot)$ utilized  during training is Transformer \cite{vaswani2017attention}, and the image encoder $I(\cdot)$ employed during inference defaults to ResNet-50 \cite{he2016deep}. Notably, both text and image encoders remain frozen throughout the entire pipeline.

\textbf{Style Generation Strategy.} We regenerate $K$ different style word vectors at the beginning of each epoch, where $K=80$. For style word vector generation, we use two distinct methods: Random and StyleMix. In `Random' generation method, we sample the style word vector from one of the five distributions (\textit{ie}, normal, xavier uniform, xavier normal, kaiming normal and kaiming uniform). When employing `StyleMix' method, we take a weighted sum of $L$ predefined word vectors to generate a new style word vector. Here, $L=8$, and the weighting coefficients ($\lambda$) are sampled from a $Beta(\alpha, \alpha)$ distribution, where $\alpha=0.1$ \cite{zhou2020domain}. At each epoch's style refresh, we randomly select one of the methods with a $50\%$ probability for regenerating styles.

\textbf{Training Style Remover and Classifier.} We train the style remover and classifier for $100$ epochs using the SGD optimizer with a learning rate of $0.008$, a momentum of $0.9$, and a batch size of $128$. Importantly, for model ensemble, we employ three different initial text templates when training the style remover and classifier. Furthermore, we utilize the ArcFace \cite{deng2019arcface} loss function, configuring it with a scaling factor of $5$ and an angular margin of $0.5$.

\textbf{Inference.} Input images are pre-processed in the same way with the CLIP model; resized to $224 \times 224$ and normalization. During inference, we employ a model ensemble strategy, where the output of the CLIP image encoder is simultaneously input into three style removers and classifiers trained with different initial text templates. In the end, we select the class corresponding to the highest score among these class scores as the prediction for the input image.

\subsection{Evaluations}
\begin{table*}
\centering
\caption{Comparison with SOTA domain generalization methods for per-domain top-1 accuracy on VLCS. We repeat each experiment using three different seeds, and report average accuracies with standard errors. ZS-CLIP (C) refers to zero-shot CLIP using `[class]' as its text prompt, and ZS-CLIP (PC) indicates zero-shot CLIP using `a photo of a [class]' as its text prompt. 
\textbf{Bold} indicates the best result. `*' denotes our reproduced results for PromptStyler, and `\dag' denotes that the paper corresponding to the method does not give the standard errors. }
 \resizebox{\linewidth}{!}{
\begin{tabular}{l|c|ccccc} 
\toprule
\multicolumn{1}{l|}{\multirow{2}{*}{Method}} &  \multirow{2}{*}{Venue}  & \multicolumn{5}{c}{Accuracy (\%) $\uparrow$}                                                                      \\ 
\cline{3-7}
\multicolumn{1}{c|}{}                        &                        & \quad\makebox[0.1\textwidth][c]{Caltech}              & \makebox[0.1\textwidth][c]{LabelMe}              &\makebox[0.1\textwidth][c]{SUN09}      & \makebox[0.1\textwidth][c]{VOC2007}          &  \makebox[0.1\textwidth][c]{Avg.}           \\ 
\hline
\multicolumn{7}{c}{ResNet-50 with pre-trained weights on ImageNet}                                                                                                             \\ 
\hline
SAGM$^{\dag}$ \cite{wang2023sharpness}  &  CVPR'2023  &  \quad 97.9 ± 0.0           &  65.0 ± 0.0  &  70.6 ± 0.0           &  76.3 ± 0.0           &  77.4            \\ 
SelfReg \cite{kim2021selfreg}                                      &  \makebox[0.14\textwidth][c]{ICCV'2021}             &  \quad\makebox[0.1\textwidth][c]{96.7 ± 0.4}           &  \makebox[0.1\textwidth][c]{\textbf{65.2 ± 1.2}}           &  \makebox[0.1\textwidth][c]{73.1 ± 1.3}           &  \makebox[0.1\textwidth][c]{76.2 ± 0.7}           &  \makebox[0.1\textwidth][c]{77.8}          \\
M$^2$-CL$^{\dag}$ \cite{ballas2024multi}  &  TAI'2024  &  \quad 98.2 ± 0.0           &  63.9 ± 0.0  &  72.4 ± 0.0           &  76.7 ± 0.0           &  77.8           \\ 
Model Ratatouille$^{\dag}$ \cite{rame2023model}  & ICML'2023   & \quad\textbf{99.3 ± 0.0}           &  60.4 ± 0.0  &  73.9 ± 0.0  &  \textbf{79.5 ± 0.0}  &  78.3   \\ 
GVRT \cite{min2022grounding}                                        &  ECCV'2022             &  \quad 98.8 ± 0.1           &  64.0 ± 0.3  &  \textbf{75.2 ± 0.5}           &  77.9 ± 1.0           &  \textbf{79.0}            \\

\hline
\multicolumn{7}{c}{ResNet-50 with pre-trained weights from CLIP}                                                                                                               \\ 
\hline
ZS-CLIP (C) \cite{radford2021learning}                                 & ICML'2021              &  \quad 99.5 ± 0.0           &  67.8 ± 0.0           &  69.5 ± 0.0           &  80.8 ± 0.0           &  79.4            \\
ZS-CLIP (PC) \cite{radford2021learning}                                 &  ICML'2021              &  \quad 99.8 ± 0.0           &  69.6 ± 0.0           &  71.0 ± 0.0           &  87.7 ± 0.0           &  82.0           \\ 
PromptStyler* \cite{cho2023promptstyler}                              &  ICCV'2023              &  \quad \textbf{100.0 ± 0.0}         & 72.5 ± 1.1           &  67.9 ± 1.3           &  88.4 ± 0.6           &  82.2            \\ \hline
\rowcolor[rgb]{0.929,0.929,0.929} DPStyler   &  Ours                  &  \quad\textbf{100.0 ± 0.0}   &  \textbf{73.3 ± 0.9}  &  \textbf{71.8 ± 0.9}  &  \textbf{89.0 ± 0.1}  & \makebox[0.1\textwidth][c]{$\textbf{83.5}$}   \\ 
\hline
\multicolumn{7}{c}{ViT-B / 16 with pre-trained weights from CLIP}                                                                                                              \\ 
\hline
ZS-CLIP (C) \cite{radford2021learning}                                  & ICML'2021            & \quad 99.8 ± 0.0           &  60.9 ± 0.0           &   69.8 ± 0.0           &     74.1 ± 0.0       &  76.2            \\
ZS-CLIP (PC) \cite{radford2021learning}                                 &  ICML'2021              &  \quad \textbf{100.0 ± 0.0}           &  70.0 ± 0.0           &  74.1 ± 0.0           &    88.0 ± 0.0         &  83.0            \\
PromptStyler* \cite{cho2023promptstyler}                                &  ICCV'2023              & \quad \textbf{100.0 ± 0.0}           &  \textbf{72.5 ± 0.4}  & 72.4 ± 1.9          &  89.9 ± 0.3           &  83.7            \\ \hline
\rowcolor[rgb]{0.929,0.929,0.929} DPStyler   & Ours                  &  \quad\textbf{100.0 ± 0.0}  &  69.1 ± 1.3  &  \textbf{76.3 ± 0.2}  &  \textbf{90.5 ± 0.4}  &  \makebox[0.1\textwidth][c]{$\textbf{84.0}$}   \\ 
\hline
\multicolumn{7}{c}{ViT-L / 14 with pre-trained weights from CLIP}                                                                                                              \\ 
\hline
ZS-CLIP (C) \cite{radford2021learning}                                  & ICML'2021              & \quad \textbf{100.0 ± 0.0}           &  57.5 ± 0.0           &  70.5 ± 0.0           &   82.1 ± 0.0          &  77.5            \\
ZS-CLIP (PC) \cite{radford2021learning}                                 &  ICML'2021             &  \quad \textbf{100.0 ± 0.0}           &  \textbf{70.8 ± 0.0}           &    68.6 ± 0.0         &     \textbf{88.1 ± 0.0}        &  81.9            \\
PromptStyler* \cite{cho2023promptstyler}                                &  ICCV'2023              & \quad \textbf{100.0 ± 0.0}  &  66.3 ± 1.1           &  71.8 ± 2.2           &  87.0 ± 0.5                    &  81.3              \\ \hline
\rowcolor[rgb]{0.929,0.929,0.929} DPStyler   & Ours                   &  \quad \textbf{100.0 ± 0.0}  &  68.2 ± 0.4  & \textbf{76.4 ± 0.5}  &  \textbf{88.1 ± 0.3}                    &  \makebox[0.1\textwidth][c]{$\textbf{83.2}$}              \\
\bottomrule
\end{tabular}
 }
\label{tab:table_vlcs}
\vspace{-0.2cm}
\end{table*}

\label{sec:4.3}
\begin{figure}
    \centering
    \includegraphics[width=0.95\columnwidth]{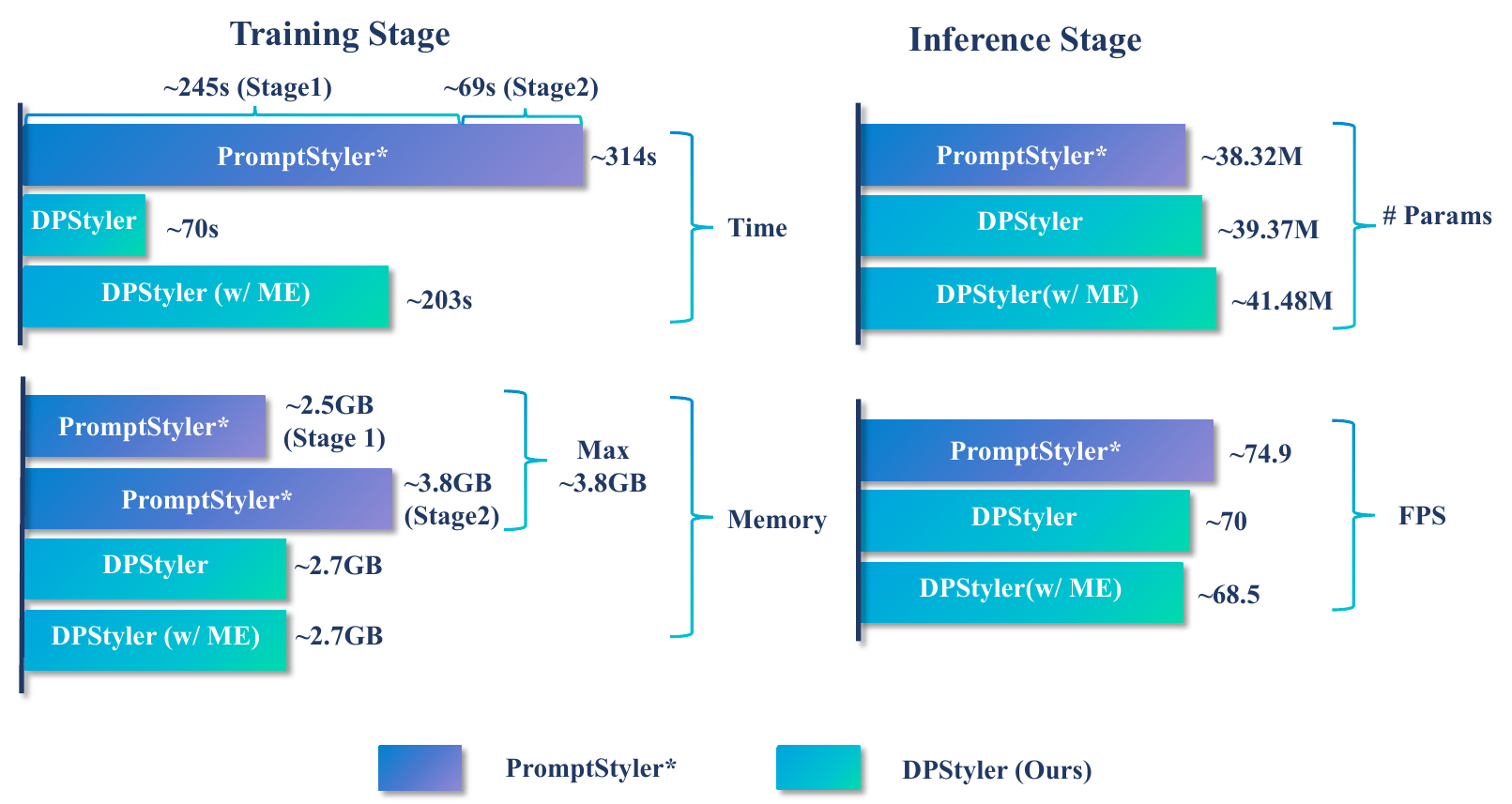}
    \caption{The evaluation of training and inference resources on VLCS including GPU memory usage, training time, model parameter count, and inference speed. Stage 1 represents the training of style word vectors by PromptStyler, and stage 2 denotes the training of the classifier. The symbol `*' denotes reproduced results. `w/ ME' stands for using model ensemble.}
    \label{figure_resource}
    \vspace{-0.3cm}
\end{figure}

\begin{figure}
    \centering
    \subcaptionbox{Before the style remover \label{fig:t_sne_a}}
      {\includegraphics[width=0.48\linewidth]{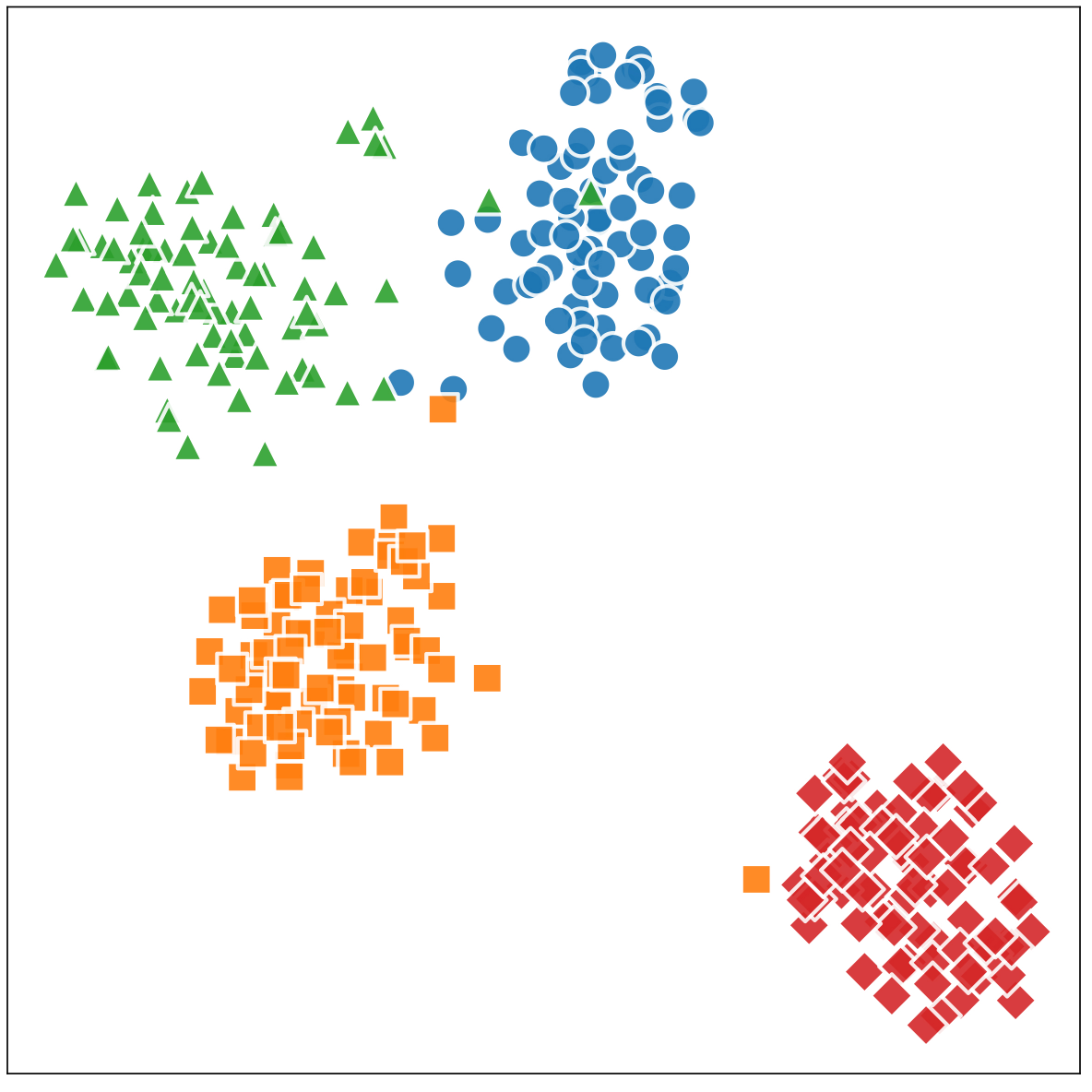}}
    \subcaptionbox{After the style remover\label{fig:t_sne_b}}
      {\includegraphics[width=0.48\linewidth]{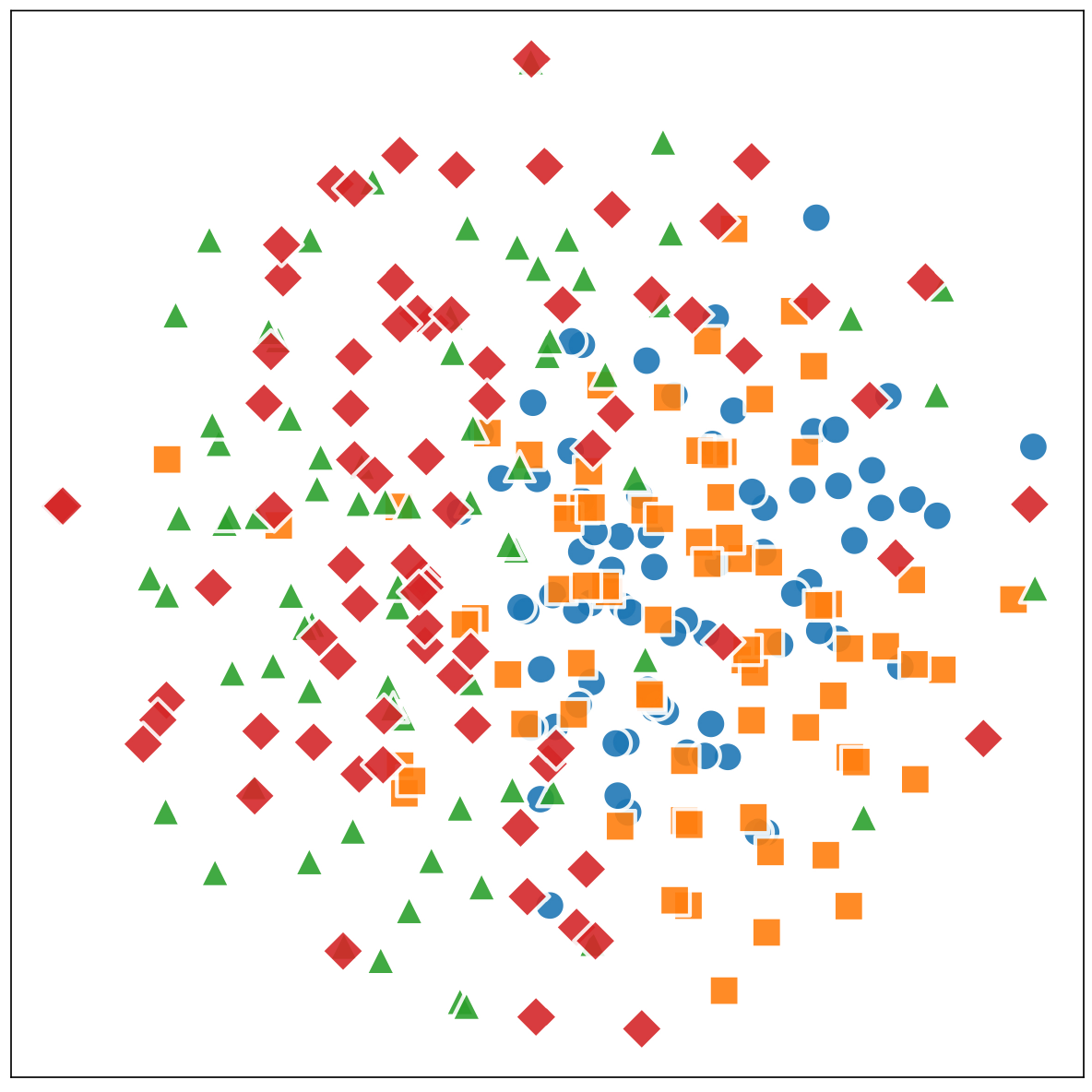}}
    \caption{The t-SNE \cite{t_sne} visualization results show the features extracted by the CLIP image encoder. These features are for the same category (\textit{e.g.}, `dog') before and after applying the style remover on the PACS dataset. Different colors represent distinct domains within PACS (\textit{e.g.}, `photo', `sketch').}
    \label{fig:t_sne}
    \vspace{-0.39cm}
\end{figure}

\textbf{Main Results.} 
DPStyler establishes the state-of-the-art performance on each dataset, as demonstrated in Table \ref{tab:table1}.
We have conducted a comprehensive comparative analysis of our method against the PromptStyler and other methods that leverage source domain data. Additionally, we have presented the performance of zero-shot CLIP, which generates each text feature using a domain-agnostic prompt (``[class]"), denoted as ZS-CLIP (C) and a domain-specific prompts (``a photo of a [class]"), denoted as ZS-CLIP (PC). Note that all existing methods utilize source domain data except for zero-shot CLIP, PromptStyler \cite{cho2023promptstyler} and ours in Table \ref{tab:table1}. Remarkably, our approach consistently outperforms other methods across all datasets. 
As illustrated in Table \ref{tab:table1}, DPStyler achieves comparable or even superior performance to PromptStyler while utilizing fewer training resources. 

\textbf{Per-domain Accuracy.} As shown in Table \ref{tab:table_pacs}-\ref{tab:table_domainnet}, we provide per-domain top-1 classification accuracy on four domain generalization benchmarks: PACS, VLCS, OfficeHome, and DomainNet. The accuracy is calculated by averaging results across experiments conducted with three different random seeds. It can be seen that our method performs well on each domain of each dataset.

\begin{table*}
\centering
\caption{Comparison with SOTA domain generalization methods for per-domain top-1 accuracy on OfficeHome. We repeat each experiment using three different seeds, and report average accuracies with standard errors. ZS-CLIP (C) refers to zero-shot CLIP using `[class]' as its text prompt, and ZS-CLIP (PC) indicates zero-shot CLIP using `a photo of a [class]' as its text prompt. 
\textbf{Bold} indicates the best result. `*' denotes our reproduced results for PromptStyler, and `\dag' denotes that the paper corresponding to the method does not give the standard errors. }
 \resizebox{\linewidth}{!}{
\begin{tabular}{l|c|ccccc} 
\toprule
\multicolumn{1}{l|}{\multirow{2}{*}{Method}} &  \multirow{2}{*}{Venue}  & \multicolumn{5}{c}{Accuracy (\%) $\uparrow$}                                                                      \\ 
\cline{3-7}
\multicolumn{1}{c|}{}                        &                        & \quad\makebox[0.1\textwidth][c]{Art}              & \makebox[0.1\textwidth][c]{Clipart}              &\makebox[0.1\textwidth][c]{Product}      & \makebox[0.1\textwidth][c]{Real World}          &  \makebox[0.1\textwidth][c]{Avg.}           \\ 
\hline
\multicolumn{7}{c}{ResNet-50 with pre-trained weights on ImageNet}                                                                                                             \\ 
\hline

SelfReg \cite{kim2021selfreg}                                      &  \makebox[0.14\textwidth][c]{ICCV'2021}             &  \quad\makebox[0.1\textwidth][c]{63.6 ± 1.4}           &  \makebox[0.1\textwidth][c]{53.1 ± 1.0}           &  \makebox[0.1\textwidth][c]{76.9 ± 0.4}           &  \makebox[0.1\textwidth][c]{78.1 ± 0.4}           &  \makebox[0.1\textwidth][c]{67.9}          \\

SAGM$^{\dag}$ \cite{wang2023sharpness}  &  CVPR'2023  &  \quad 63.3 ± 0.0           &  54.3 ± 0.0  &  76.4 ± 0.0           &  77.8 ± 0.0           &  67.9            \\ 
GVRT \cite{min2022grounding}                                        &  ECCV'2022             &  \quad 66.3 ± 0.1           &  55.8 ± 0.4  &  78.2 ± 0.4           &  \textbf{80.4 ± 0.2}           &  70.1            \\ 
M$^2$-CL$^{\dag}$ \cite{ballas2024multi}  &  TAI'2024  &  \quad \textbf{68.5 ± 0.0}           &  \textbf{57.0 ± 0.0}  &  \textbf{78.7 ± 0.0}           &  80.1 ± 0.0           &  \textbf{71.1}           \\ 

\hline
\multicolumn{7}{c}{ResNet-50 with pre-trained weights from CLIP}                                                                                                               \\ 
\hline
ZS-CLIP (C) \cite{radford2021learning}                                 & ICML'2021              &  \quad 68.7 ± 0.0           &  44.4 ± 0.0           &  77.1 ± 0.0           &  79.5 ± 0.0           &  67.4            \\
ZS-CLIP (PC) \cite{radford2021learning}                                 &  ICML'2021              &  \quad 71.1 ± 0.0           &  50.0 ± 0.0           &  81.3 ± 0.0           &  82.0 ± 0.0           &  71.1           \\ 
PromptStyler* \cite{cho2023promptstyler}                              &  ICCV'2023                   & \quad 71.3 ± 0.5 & 48.8 ± 0.5           &  81.9 ± 0.6           &  82.2 ± 0.2           &  71.1            \\ \hline
\rowcolor[rgb]{0.929,0.929,0.929} DPStyler   &  Ours                  &  \quad\textbf{73.6 ± 0.4}   &  \textbf{51.2 ± 0.2}  &  \textbf{82.1 ± 0.6}  &  \textbf{82.9 ± 0.4}  & \makebox[0.1\textwidth][c]{$\textbf{72.5}$}   \\ 
\hline
\multicolumn{7}{c}{ViT-B / 16 with pre-trained weights from CLIP}                                                                                                              \\ 
\hline
ZS-CLIP (C) \cite{radford2021learning}                                  & ICML'2021            & \quad 80.9 ± 0.0           &  64.3 ± 0.0           &   85.9 ± 0.0           &     87.2 ± 0.0       &  79.6            \\
ZS-CLIP (PC) \cite{radford2021learning}                                 &  ICML'2021              &  \quad 83.1 ± 0.0           &  65.8 ± 0.0           &  89.1 ± 0.0           &    89.2 ± 0.0         &  81.8            \\
PromptStyler* \cite{cho2023promptstyler}                                &  ICCV'2023              & \quad 81.8 ± 0.6           &  66.0 ± 1.0  & 89.7 ± 0.4          &  89.6 ± 0.1           &  81.8            \\ \hline
\rowcolor[rgb]{0.929,0.929,0.929} DPStyler   & Ours                  &  \quad\textbf{83.6 ± 0.4}  &  \textbf{67.8 ± 0.5}  &  \textbf{90.0 ± 0.2}  &  \textbf{89.7 ± 0.1}  &  \makebox[0.1\textwidth][c]{$\textbf{82.8}$}   \\ 
\hline
\multicolumn{7}{c}{ViT-L / 14 with pre-trained weights from CLIP}                                                                                                              \\ 
\hline
ZS-CLIP (C) \cite{radford2021learning}                                  & ICML'2021              & \quad 86.4 ± 0.0           &  72.3 ± 0.0           &  92.3 ± 0.0           &   91.8 ± 0.0          &  85.7            \\
ZS-CLIP (PC) \cite{radford2021learning}                                 &  ICML'2021             &  \quad 86.8 ± 0.0           &  73.6 ± 0.0          &    92.9 ± 0.0         &    93.2 ± 0.0        &  86.6            \\
PromptStyler* \cite{cho2023promptstyler}                                &  ICCV'2023              & \quad 86.3 ± 0.3  &  73.5 ± 0.7           &  93.5 ± 0.7           &  92.2 ± 0.3                    &  86.4              \\ \hline
\rowcolor[rgb]{0.929,0.929,0.929} DPStyler   & Ours                   &  \quad \textbf{88.0 ± 0.4}  &  \textbf{76.4 ± 0.3}  & \textbf{94.3 ± 0.5}  &  \textbf{93.4 ± 0.3}                    &  \makebox[0.1\textwidth][c]{$\textbf{88.0}$}              \\
\bottomrule
\end{tabular}
 }

\label{tab:table_office}

\end{table*}

\begin{table*}
\centering
\caption{Comparison with SOTA domain generalization methods for per-domain top-1 accuracy on DomainNet. We repeat each experiment using three different seeds, and report average accuracies with standard errors. ZS-CLIP (C) refers to zero-shot CLIP using `[class]' as its text prompt, and ZS-CLIP (PC) indicates zero-shot CLIP using `a photo of a [class]' as its text prompt. 
\textbf{Bold} indicates the best result. `*' denotes our reproduced results for PromptStyler, and `\dag' denotes that the paper corresponding to the method does not give the standard errors. }
 \resizebox{\linewidth}{!}{
\begin{tabular}{l|c|ccccccc} 
\toprule
\multicolumn{1}{l|}{\multirow{2}{*}{Method}} &  \multirow{2}{*}{Venue}  & \multicolumn{5}{c}{Accuracy (\%) $\uparrow$}                                                                      \\ 
\cline{3-9}
\multicolumn{1}{c|}{}                        &                        & \quad\makebox[0.1\textwidth][c]{Clipart}              & \makebox[0.1\textwidth][c]{Infograph}              &\makebox[0.1\textwidth][c]{Painting}      & \makebox[0.1\textwidth][c]{Quickdraw}         &\makebox[0.1\textwidth][c]{Real} &\makebox[0.1\textwidth][c]{Sketch} &  \makebox[0.1\textwidth][c]{Avg.}           \\ 
\hline
\multicolumn{7}{c}{ResNet-50 with pre-trained weights on ImageNet}                                                                                                             \\ 
\hline

SelfReg \cite{kim2021selfreg}                                      &  \makebox[0.14\textwidth][c]{ICCV'2021}             &  \quad\makebox[0.1\textwidth][c]{60.7 ± 0.1}           &  \makebox[0.1\textwidth][c]{21.6 ± 0.1}           &  \makebox[0.1\textwidth][c]{49.4 ± 0.2}           &  \makebox[0.1\textwidth][c]{12.7 ± 0.1}           &  \makebox[0.1\textwidth][c]{60.7 ± 0.1}         &  \makebox[0.1\textwidth][c]{51.7 ± 0.1} &  \makebox[0.1\textwidth][c]{42.8}  \\
GVRT \cite{min2022grounding}                                        &  ECCV'2022             &  \quad 62.4 ± 0.4           &  21.0 ± 0.0  &  50.5 ± 0.4           &  13.8 ± 0.3           &  64.6 ± 0.4    &  52.4 ± 0.2         &  44.1  \\ 
Model Ratatouille$^{\dag}$ \cite{rame2023model}  & ICML'2023   & \quad\textbf{66.1 ± 0.0}           &  \textbf{23.1 ± 0.0}  &   \textbf{55.5 ± 0.0}  &  \textbf{16.7 ± 0.0}  & \textbf{68.5 ± 0.0}&\textbf{56.0 ± 0.0}& \textbf{47.7}   \\ 

\hline
\multicolumn{7}{c}{ResNet-50 with pre-trained weights from CLIP}                                                                                                               \\ 
\hline
ZS-CLIP (C) \cite{radford2021learning}                                 & ICML'2021              &  \quad 52.8 ± 0.0           &  40.1 ± 0.0           &  52.9 ± 0.0           &  6.5 ± 0.0   &  75.3 ± 0.0   &  47.6 ± 0.0           &  45.9            \\
ZS-CLIP (PC) \cite{radford2021learning}                                 &  ICML'2021              &  \quad 53.1 ± 0.0           &  39.6 ± 0.0           &  52.7 ± 0.0           &  5.6 ± 0.0   &  76.8 ± 0.0 &  48.5 ± 0.0           &  46.1           \\ 
PromptStyler* \cite{cho2023promptstyler}                              &  ICCV'2023                   & \quad 53.9 ± 0.7 & 41.4 ± 0.1           &  54.6 ± 0.6           &  5.6 ± 0.3    &  76.8 ± 0.3    &  49.2 ± 0.2     &  46.9            \\ \hline
\rowcolor[rgb]{0.929,0.929,0.929} DPStyler   &  Ours                  &  \quad\textbf{55.4 ± 0.4}   &  \textbf{41.5 ± 0.1}  &  \textbf{56.0 ± 0.1}  &  \textbf{7.1 ± 0.2} &  \textbf{77.5 ± 0.0}  &  \textbf{50.7 ± 0.1} & \makebox[0.1\textwidth][c]{$\textbf{48.0}$}   \\ 

\hline
\multicolumn{7}{c}{ViT-B / 16 with pre-trained weights from CLIP}                                                                                                              \\ 
\hline
ZS-CLIP (C) \cite{radford2021learning}                                  & ICML'2021            & \quad 70.2 ± 0.0           &  48.9 ± 0.0           &   65.7 ± 0.0           &     14.3 ± 0.0     &     82.4 ± 0.0  &     62.7 ± 0.0    &  57.4            \\
ZS-CLIP (PC) \cite{radford2021learning}                                 &  ICML'2021         &    \quad  70.4 ± 0.0        & 47.3 ± 0.0           &  65.0 ± 0.0           &  13.5 ± 0.0           &    83.3 ± 0.0    &    63.6 ± 0.0     &  57.2           \\
PromptStyler* \cite{cho2023promptstyler}                                &  ICCV'2023               &    \quad  70.1 ± 0.5        & 47.7 ± 0.7           &  65.1 ± 0.3           &  12.5 ± 0.6           &    82.3 ± 0.3    &    62.3 ± 0.5     &  56.7           \\ \hline
\rowcolor[rgb]{0.929,0.929,0.929} DPStyler   & Ours                  &  \quad\textbf{71.5 ± 0.2}  &  \textbf{50.6 ± 0.1}  &  \textbf{66.8 ± 0.3}  &  \textbf{16.2 ± 0.1} &  \textbf{83.7 ± 0.0}   &  \textbf{64.4 ± 0.1} &  \makebox[0.1\textwidth][c]{ $\textbf{58.9}$}   \\ 
\hline
\multicolumn{7}{c}{ViT-L / 14 with pre-trained weights from CLIP}                                                                                                              \\ 
\hline
ZS-CLIP (C) \cite{radford2021learning}                                  & ICML'2021              & \quad 77.6 ± 0.0           &  52.7 ± 0.0           &  71.0 ± 0.0           &   21.6 ± 0.0   &   85.9 ± 0.0    &   70.0 ± 0.0      &  63.1           \\
ZS-CLIP (PC) \cite{radford2021learning}                                 &  ICML'2021            & \quad \textbf{78.3 ± 0.0}           &  50.6 ± 0.0           &  69.0 ± 0.0           &   22.4 ± 0.0   &   86.3 ± 0.0    &   71.5 ± 0.0      &  63.0           \\
PromptStyler* \cite{cho2023promptstyler}                                &  ICCV'2023              & \quad 77.5 ± 0.5           &  52.3 ± 1.3           &  70.8 ± 1.0           &   21.0 ± 0.7   &   86.1 ± 0.2    &   69.5 ± 0.8      &  62.9           \\ \hline
\rowcolor[rgb]{0.929,0.929,0.929} DPStyler   & Ours                   &  \quad \textbf{78.3 ± 0.1}  &  \textbf{54.6 ± 0.0}  & \textbf{73.2 ± 0.1}  &  \textbf{23.7 ± 0.2}      &  \textbf{86.7 ± 0.0}       &  \textbf{71.9 ± 0.2}         &  \makebox[0.1\textwidth][c]{$\textbf{64.7}$}              \\
\bottomrule
\end{tabular}
 }

\label{tab:table_domainnet}
\vspace{-0.2cm}
\end{table*}

\textbf{Computational Evaluations.} 
In Figure \ref{figure_resource}, we conduct a resource usage comparison between DPStyler (ours) and PromptStyler on the VLCS dataset, encompassing both training and inference resources. 
During the training stage, PromptStyler requires a two-stage training process referred to as `stage 1' and `stage 2', with $5$ minutes for total training. In contrast, our approach significantly reduces the demand for training resources. Utilizing two methods for generating style word vectors without additional training resources results in a requirement of only $20\%$ of PromptStyler's training time and $70\%$ of its GPU memory usage when excluding model ensemble method. Even when model ensemble is used, our method only requires $60\%$ of the training time compared to PromptStyler, while keeping GPU memory requirements consistent with those without using model ensemble.
Finally, during the inference stage, we conduct experiments with a batch size of $1$. As shown in Figure \ref{figure_resource}, our method slightly increases the model parameter count compared to PromptStyler. However, due to the minimal increase in parameter count, our inference speed remains very close to that of PromptStyler, whether or not model ensemble is utilized.
\begin{figure}[h]
    \centering
    \includegraphics[width=1\columnwidth]{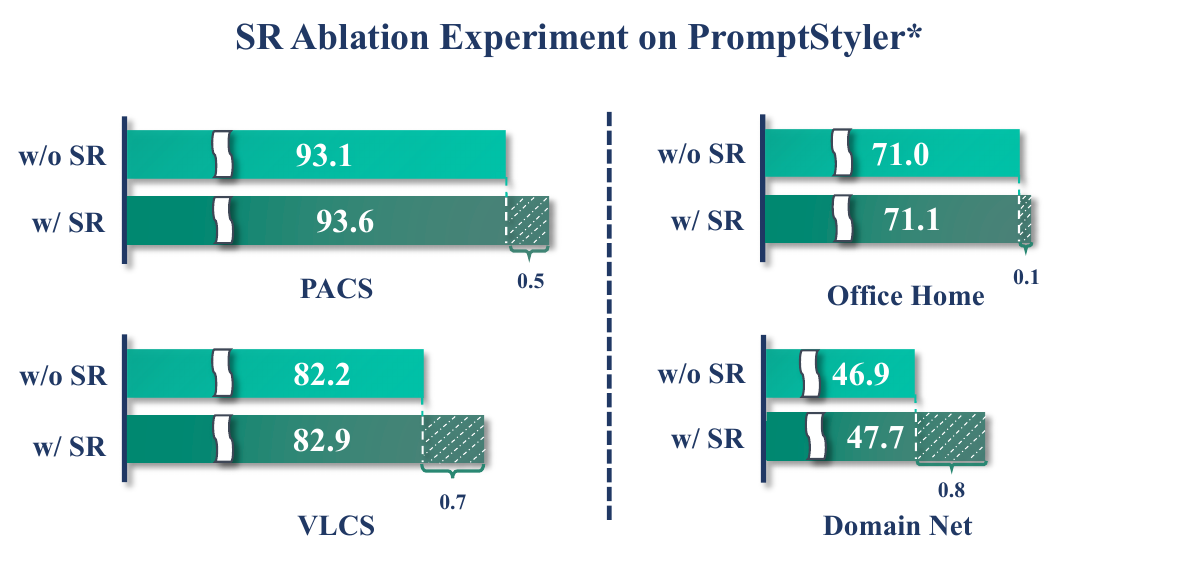}
    \vspace{-0.5cm}
    \caption{Ablation experiments to assess the effectiveness of Style Remover (SR) within PromptStyler across PACS, VLCS, OfficeHome, and DomainNet. `w/' and `w/o' respectively indicate whether Style Remover is used in PromptStyler.}
    \label{figure_ps_sr}
    \vspace{-0.1cm}
\end{figure}

\textbf{t-SNE Visualization Results.} In Figure \ref{fig:t_sne}, we employ t-SNE \cite{t_sne} to visualize the features before and after style remover on the PACS dataset. Figure \ref{fig:t_sne_a} displays notable feature variations within the CLIP image encoder output for the same category (\textit{e.g.}, 'dog'), suggesting style information. However, Figure \ref{fig:t_sne_b} demonstrates that after style remover, these features share similar semantics, indicating successful elimination of style information. These results prove the effectiveness of our style removal module in eliminating style  features from CLIP's generated features.
\vspace{-0.1cm}

\subsection{Ablation Study}

\begin{table}[]
\centering
\caption{Ablation studies on Style Generation (SG), Style Remover (SR), and Model Ensemble (ME) across PACS, VLCS, OfficeHome (OH), and DomainNet (DN).}
\resizebox{\linewidth}{!}{
    \begin{tabular}{ccc|cccc|c} 
    \toprule
    \multicolumn{3}{c|}{Component}              & \multicolumn{5}{c}{Accuracy (\%) $\uparrow$}               \\ 
    \hline
      SG               & SR            & ME            & PACS       &  VLCS         &  OH           & DN        & Avg.   \\ \hline
      -                & -             & -               & 91.7     & 76.8          & 70.0          & 44.9      &70.9        \\
     $\checkmark$      & -             & -              &   93.3     &     83.2       &     70.7       &    47.5   & 73.7   \\
     $\checkmark$      & $\checkmark$  & -              &  \textbf{93.9}      & 83.2          & 71.5  &   47.3    & 74.0  \\
      $\checkmark$      & -            & $\checkmark$   & 93.1      & \textbf{83.6} & 71.9      & \textbf{48.0}     & 74.1   \\
    \rowcolor[rgb]{0.929,0.929,0.929} $\checkmark$      & $\checkmark$  & $\checkmark$   & 93.6 & 83.5      & \textbf{72.5}          & \textbf{48.0}   & \textbf{74.4}  \\ 
    \bottomrule
    \end{tabular}
}

\label{tab:table3}
\vspace{-0.4cm}
\end{table}

We conduct experiments on the three distinct components of our model to assess their specific contributions to the overall performance within our proposed framework.
Notably, we train a new baseline model that omits these three modules, resulting in a model without these components. This baseline model is trained with a single classifier, where the input during the training phase is the class name of each dataset. The training and inference processes align with the methods outlined in Section \ref{subsec:4.2}. Note that we use the text template ``a [class] in a $\boldsymbol{S}_*$ style'' in all ablation experiments except when testing the model ensemble, three different text templates are used to train distinct classifiers. The results of the ablation study are presented in Table \ref{tab:table3}.

\begin{figure}
    \centering
    \includegraphics[width=1\columnwidth]{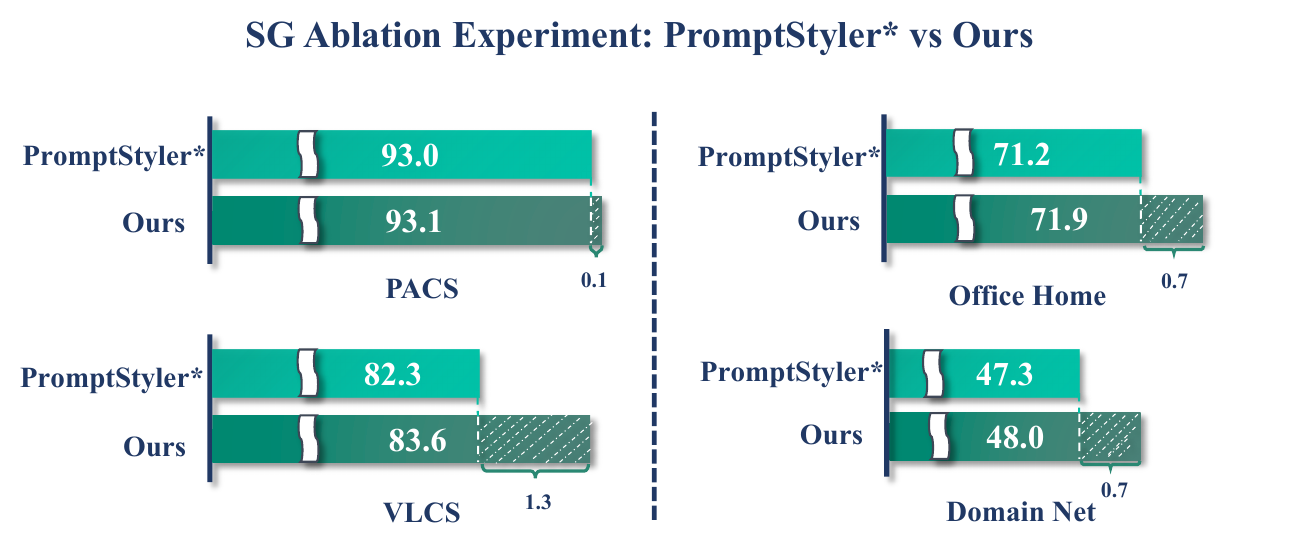}
    \vspace{-0.5cm}
     \caption{Ablation experiments to compare the effect of promptstyler with the Style Generation Module (SG) in our method on four benchmark datasets. To ensure a fair comparison, we apply the model ensemble method to both PromptStyler and DPStyler simultaneously. `*' denotes reproduced results.}
    \label{figure_sg}
    \vspace{-0.07cm}
\end{figure}

\textbf{Style Generation.} We conduct experiments to evaluate DPStyler's style generation module. As shown in the second row of Table \ref{tab:table3}, it's evident that our classifier's performance has been improved by employing the style generation module alone. This suggests that our style generation module effectively provides more diverse features to the classifier, thereby enhancing its robustness and generalization ability. Furthermore, in order to show that dynamic update styles make the model performs better than fixed styles, we compare the performances of PromptStyler and DPStyler on the four datasets without using style remover and with model ensemble for both. As can be seen from Figure \ref{figure_sg}, our dynamic style generation module is able to obtain better results than PromptStyler with less training resources. Additionally, we conduct separate tests for using Random only, using StyleMix only, and Random-Mix ($50\%$ Random, $50\%$ StyleMix), as shown in Table \ref{tab:table4}. 

Additionally, we extend our comparison of word vector generation methods to include the 'Gaussian' strategy, where style word vectors are sampled from a zero-mean Gaussian distribution with a $0.02$ standard deviation. Different from other prompt learning methods \cite{zhou2022conditional}, style word vectors are resampled at each training epoch in `Gaussian' for fair comparison. Experimental results indicate that solely sampling style from 'Gaussian' can achieve performance comparable to that of promptStyler. Furthermore, the 'Random-Mix' generation strategy yields the best overall performance.

\begin{table}[]
\centering
\caption{Ablation experiments on different model heads across PACS, VLCS, OfficeHome, and DomainNet. Here 'MLP' represents a fully connected layer as a style remover, and 'Style-SE Net' represents our method. }
\resizebox{\linewidth}{!}{
    \begin{tabular}{l|cccc|c} 
    \toprule
    \multirow{2}{*}{Head} & \multicolumn{5}{c}{Accuracy (\%) $\uparrow$}                                              \\ 
    \cline{2-6}
                              & PACS          & VLCS          & OfficeHome            & DomainNet            & Avg.           \\ 
    \hline
    MLP                &    93.3       & 82.5  &      71.9     &  47.2       &    73.7        \\
    \rowcolor[rgb]{0.929,0.929,0.929} Style-SE Net  & \textbf{93.6}  &    \textbf{83.5}       & \textbf{72.5}  & \textbf{48.0} &  \textbf{74.4} \\
    \bottomrule
    \end{tabular}
}
\vspace{-0.05cm}

\label{tab:table_head}
\vspace{-0.3cm}
\end{table}
\textbf{Style Remover.} As seen in Table \ref{tab:table3}'s third row, our evaluation of the integrated style removal module in DPStyler shows further enhancement in classifier performance beyond only using the style generation module.
Additionally, to further assess this style removal module's effectiveness, we apply it to PromptStyler, depicted in Figure \ref{figure_ps_sr}. We observe that PromptStyler demonstrate improved performance after the inclusion of the style removal module. 

We also explore other structures for the style removal module, as seen in Table \ref{tab:table_head}. Since the module processes the output vectors of the encoder, the structure should not be too complex. We attempted to use a fully connected layer as the style remover and compared it with the Style-SE Net proposed in this paper. It can be observed that a fully connected layer can also achieve the effect of style removal, but the performance of Style-SE Net is superior.


\begin{table}[]
\centering
\caption{
The ablation experiments on the Model Ensemble Module using three text templates across PACS, VLCS, OfficeHome (OH), and DomainNet (DN). Ensemble represents the performance of an integrated classifier trained with three different text templates.
}
\resizebox{\linewidth}{!}{
    \begin{tabular}{c|cccc|c}
    \toprule
    
     \multirow{2}{*}{Template} & \multicolumn{5}{c}{Accuracy (\%) $\uparrow$}                                              \\ 
    \cline{2-6}
                          & PACS          & VLCS          & OH            & DN            & Avg.           \\  \hline
    {\small a [class] in a $S_*$ style}                  & \textbf{93.9} & 83.2 & 71.5 & 47.3 & 74.0 \\
    {\small a $S_*$ style of a [class]}                  & 93.1 & 77.0 & \textbf{73.0} & 47.9 & 72.8 \\
    {\scriptsize a photo of a [class] with $S_*$ like style}  & 93.7 & \textbf{83.9} & 70.8 & 45.7 & 73.5 \\ \hline
   \rowcolor[rgb]{0.929,0.929,0.929} Ensemble                                            & 93.6 & 83.5 & 72.5 & \textbf{48.0} & \textbf{74.4} \\
    \bottomrule
    \end{tabular}
}

\label{tab:table7}
\vspace{-0.1cm}
\end{table}
\begin{table}[t]
\centering
\caption{Ablation experiments on style refresh strategies across PACS, VLCS, OfficeHome (OH), and DomainNet (DN). }
\resizebox{\linewidth}{!}{
    \begin{tabular}{l|cccc|c} 
    \toprule
    \multirow{2}{*}{Strategy} & \multicolumn{5}{c}{Accuracy (\%) $\uparrow$}                                              \\ 
    \cline{2-6}
                              & PACS          & VLCS          & OH            & DN            & Avg.           \\ 
    \hline
    Gaussian     & 92.7          & 82.4          & \textbf{70.7} & 46.9          & 73.2           \\
    \hline
    Random                    &   92.9       &  82.8        &   \textbf{70.7}        &    \textbf{47.5}       &   73.5         \\
    StyleMix                  &    \textbf{93.3}       & 82.5 &     69.4     &  46.2        &    72.9        \\
    \rowcolor[rgb]{0.929,0.929,0.929} Random-Mix  & \textbf{93.3}  &    \textbf{83.2}       & \textbf{70.7}  & \textbf{47.5} &  \textbf{73.7} \\
    \bottomrule
    \end{tabular}
}
\vspace{-0.1cm}

\label{tab:table4}
\vspace{-0.2cm}
\end{table}

\textbf{Model Ensemble.} In Table \ref{tab:table3}, the fourth and fifth rows illustrate the performance evaluation of an integrated classifier trained using three different text templates. The table demonstrates a significant improvement in overall model performance through model ensemble.
Furthermore, Table \ref{tab:table7} highlights that while integrating classifiers trained with multiple text templates may not yield optimal performance for every dataset, it notably enhances overall performance stability. This enhancement reduces the model's susceptibility to significant fluctuations resulting from variations in text templates. 

We show the impact of different model ensemble methods on the results. In this paper, we select the class corresponding to the maximum value from the class scores generated by all models as the prediction. However, model ensemble can also be achieved through averaging. That is, for the class scores generated by all models, the scores for the same class are summed and divided by the number of models to obtain the average scores for each class. Subsequently, the class corresponding to the maximum value in the average scores is selected as the prediction. In Table \ref{tab:table_ensemble} we compare the results obtained by these two different ensemble methods. It can be found that the `Max' method obtains better results than `Average' in general. On the specific datasets, both methods give almost identical results on PACS, OfficeHome and DomainNet, while on VLCS `Max' gives significantly better results than `Average'.

\begin{figure}[]
    \centering
    \vspace{-0.3cm}
    \includegraphics[width=1\columnwidth]{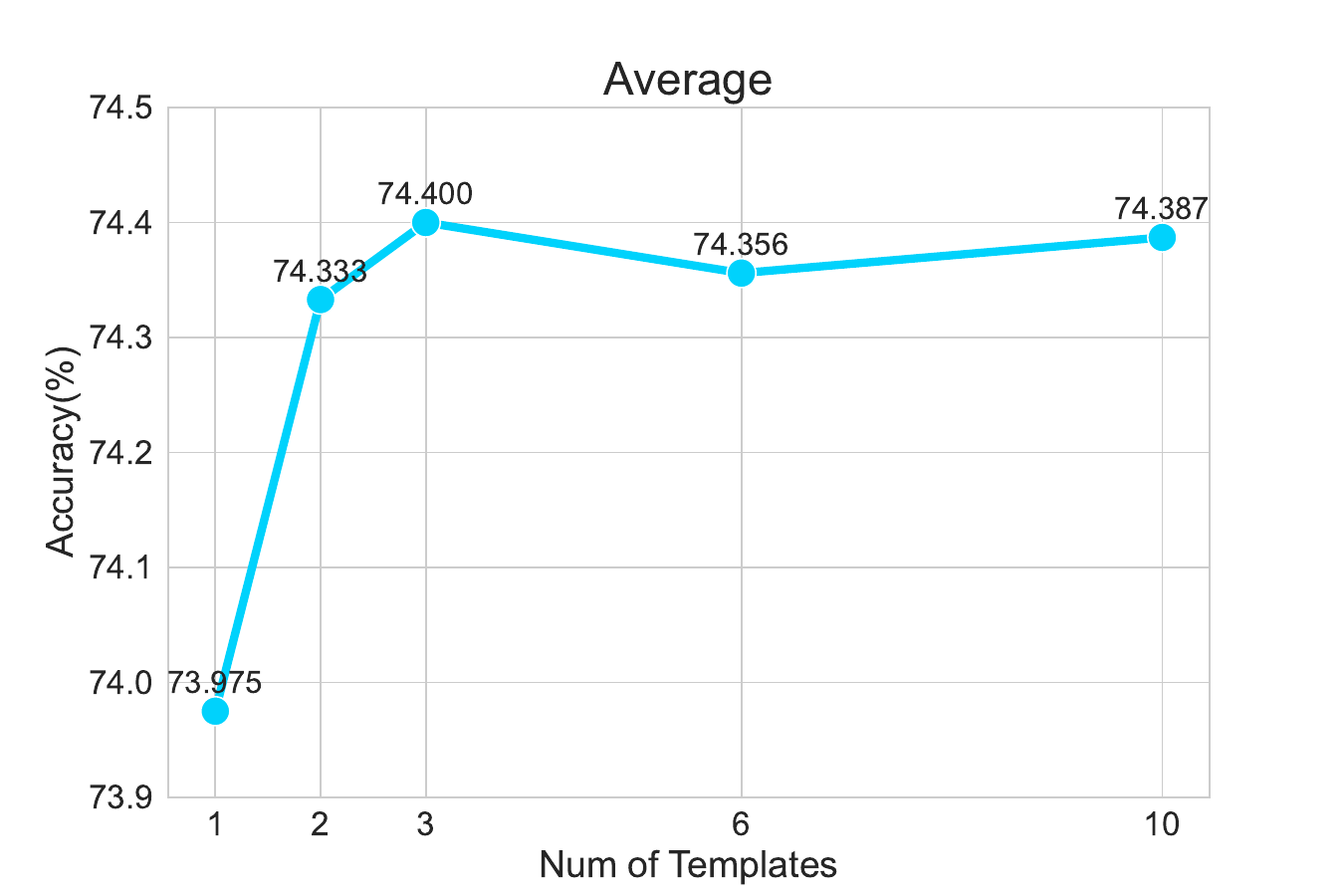}
    \caption{The average Top-1 classification accuracy on the PACS, VLCS, OfficeHome and DomainNet datasets with regard to the number of templates. When considering both model performance and resource usage, we set the number of templates to $3$.}
    \label{figure_average_acc}
    \vspace{-0.5cm}
\end{figure}

Moreover, we explore the impact of the number of templates on the effectiveness of the model. Figure \ref{figure_average_acc} illustrates the change in the average accuracy over the four domains with an increase in the number of templates. It can be found that the average accuracy increases monotonically when the number of templates increases from $1$ to $3$. As the number of templates continues to increase, the model's effectiveness decreases slightly but stays at a stable level. Therefore, it is more reasonable to set the number of templates to $3$ when considering the model performance and the resource usage.

\begin{table}[]
\centering
\caption{Ablation experiments on different model ensemble methods across PACS, VLCS, OfficeHome, and DomainNet. }
\resizebox{\linewidth}{!}{
    \begin{tabular}{l|cccc|c} 
    \toprule
    \multirow{2}{*}{Method} & \multicolumn{5}{c}{Accuracy (\%) $\uparrow$}                                              \\ 
    \cline{2-6}
                              & PACS          & VLCS          & OfficeHome            & DomainNet            & Avg.           \\ 
    \hline
    Average                &    93.5       & 82.3  &     \textbf{72.5}     &  \textbf{48.1}       &    74.1        \\
    \rowcolor[rgb]{0.929,0.929,0.929} Max  & \textbf{93.6}  &    \textbf{83.5}       & \textbf{72.5}  & 48.0 &  \textbf{74.4} \\
    \bottomrule
    \end{tabular}
}
\vspace{-0.1cm}

\label{tab:table_ensemble}
\end{table}

\begin{table*}[h]
\centering
\caption{The evaluation of training and inference resources on PACS, VLCS and OfficeHome encompasses GPU memory usage, training time, model parameter count, and inference speed. Training Stage 1 represents the training of style word vectors by PromptStyler, and Training Stage 2 denotes the training of the classifier. Training Total represents the cumulative cost of both stages. The symbol `*' denotes reproduced results. `ME' stands for model ensemble, while `w/' and `w/o' respectively indicate whether model ensemble is used.}
\resizebox{\linewidth}{!}{
\begin{tabular}{l|cc|cc|cc|cc} 
\toprule

\multirow{2}{*}{Method} & \multicolumn{2}{c!{\color{black}\vrule}}{\makebox[0.175\textwidth][c]{Training Stage 1}} & \multicolumn{2}{c|}{\makebox[0.175\textwidth][c]{Training Stage 2}}                  & \multicolumn{2}{c|}{\makebox[0.175\textwidth][c]{Training Total}}                                          & \multicolumn{2}{c}{\makebox[0.175\textwidth][c]{Inference}}          \\ 
 \cline{2-9}
 &     Memory $\downarrow$ & Time  $\downarrow$    & Memory $\downarrow$ & Time $\downarrow$ & Memory $\downarrow$ & Time  $\downarrow$ & \# Params $\downarrow$ & FPS $\uparrow$  \\  \hline

\multicolumn{9}{c}{PACS (7 classes)}                                        \\ 
\hline
ZS-CLIP  \cite{radford2021learning}  \qquad                    & 0GB    & 0s                                                & 0GB    & 0s                                            & 0GB                                              & 0s                        & 102.01M                        & 29.2  \\
PromptStyler*  \cite{cho2023promptstyler} \qquad                & 3.8GB  & 259s    & 2.5GB  & 85s & 3.8GB                                            & 344s                      & 38.32M                         &  70.0 \\ 
\arrayrulecolor{black}\hline
DPStyler w/o ME   \qquad                 & 0GB    & 0s                                                & 2.7GB  & 74s                                           & 2.7GB                                            & 74s                       & 39.37M                         &   71.2  \\
DPStyler w/ ME    \qquad                & 0GB    & 0s                                                & 2.7GB  & 219s                                          & 2.7GB                                            & 219s & 41.49M                         &  70.6 \\
\hline

\multicolumn{9}{c}{VLCS (5 classes)}                                        \\ 
\hline

ZS-CLIP  \cite{radford2021learning}  \qquad                    & 0GB    & 0s                                                & 0GB    & 0s                                            & 0GB                                              & 0s                        & 102.01M                        & 30.7  \\
PromptStyler*  \cite{cho2023promptstyler} \qquad                & 3.8GB  & 245s    & 2.5GB  & 69s & 3.8GB                                            & 314s                      & 38.32M                         & 74.9  \\ 
\arrayrulecolor{black}\hline
DPStyler w/o ME   \qquad                 & 0GB    & 0s                                                & 2.7GB  & 70s                                           & 2.7GB                                            & 70s                       & 39.37M                         & 70.0    \\
DPStyler w/ ME    \qquad                & 0GB    & 0s                                                & 2.7GB  & 203s                                          & 2.7GB                                            & 203s & 41.48M                         & 68.5  \\
\hline

\multicolumn{9}{c}{OfficeHome (65 classes)}                                        \\ 
\hline
ZS-CLIP  \cite{radford2021learning}  \qquad                    & 0GB    & 0s                                                & 0GB    & 0s                                            & 0GB                                              & 0s                        & 102.01M                        & 10.9  \\
PromptStyler*  \cite{cho2023promptstyler} \qquad                & 4.2GB  & 679s    & 2.5GB  & 346s & 4.2GB                                            & 1025s                      & 38.91M                         & 69.1 \\ 
\arrayrulecolor{black}\hline
DPStyler w/o ME   \qquad                 & 0GB    & 0s                                                & 2.7GB  & 338s                                           & 2.7GB                                            & 338s                       & 39.43M                         &  70.2  \\
DPStyler w/ ME    \qquad                & 0GB    & 0s                                                & 2.7GB  & 1004s                                          & 2.7GB                                            & 1004s & 41.67M                         & 66.8 \\
\bottomrule
\end{tabular}
}
\vspace{-0.3cm}
\label{tab:table_resource}
\end{table*}
\vspace{-0.3cm}
\subsection{Further Analysis}
\begin{figure}[]
    \centering
    \vspace{-0.1cm}
    \includegraphics[width=0.95\columnwidth]{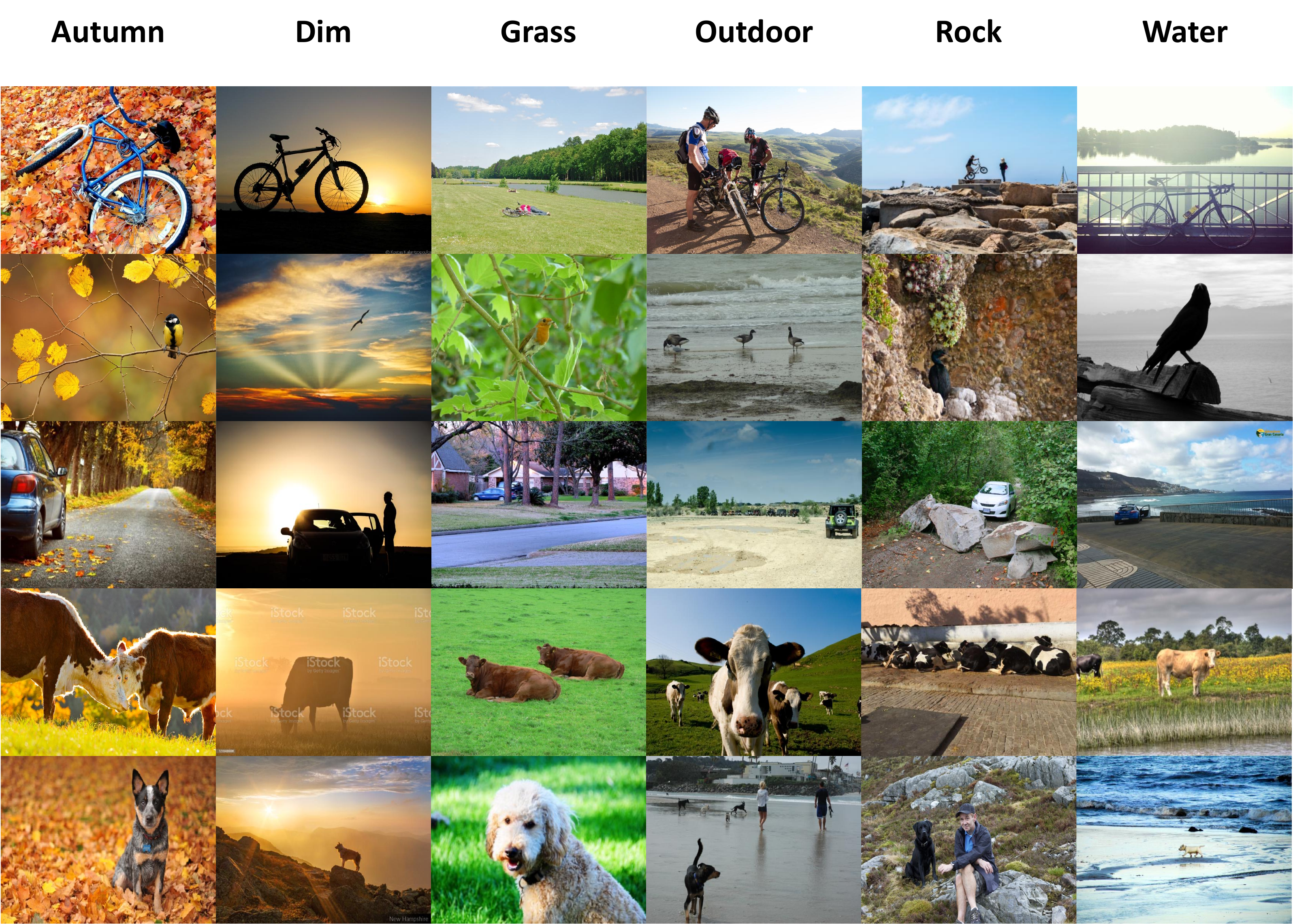}  
    \caption{Example Images of common domains in NICO++. ``Autumn", ``Dim", etc. denote domains, and we need to recognize the category of the objects in the images of each domain}
    \label{figure:nico}
    \vspace{-0.5cm}
\end{figure}

\textbf{Non-stylized Domain Shift. }When the differences between domains are primarily non-stylized, deeper-level features, we experiment with the DG benchmark on the NICO++ \cite{zhang2023nico++} dataset to verify the effectiveness of our method. Some examples of this dataset are given in Fig. \ref{figure:nico}. The images in the NICO++ dataset are all from the real world, and the differences between domains are different environments. We can see that the same object can be in a picture with grass as the background or in a picture with water as the background in Fig. \ref{figure:nico}, and that both the grass and the water belong to the non-stylized, deeper-level features. We test the performance of DPStyler on six domains on the DG benckmark \footnote{The DG benchmark for NICO++ can be found  \href{https://www.dropbox.com/sh/u2bq2xo8sbax4pr/AADbhZJAy0AAbap76cg_XkAfa?dl=0}{here}.} of this dataset and compare it with the performance of zero-shot CLIP \cite{radford2021learning}. It can be seen from TABLE \ref{tab:table_nico} that DPStyler can effectively improve the classification accuracy.

\textbf{Resource Usage. }As shown in Table \ref{tab:table_resource}, we provide detailed resource usage comparison between zero-shot CLIP \cite{radford2021learning}, PromptStyler \cite{cho2023promptstyler} and DPStyler (ours) on PACS, VLCS and OfficeHome, encompassing both training and inference resources.  From the table, it's noticeable that during the training stage without utilizing model ensemble, DPStyler's training time is only $20$\% to $30$\% of that required by PromptStyler (\textit{as depicted in Table \ref{tab:table_resource}'s training total}), with a notably significant reduction in GPU memory usage. Remarkably, under this condition, DPStyler achieves an accuracy of $74.0$\%, surpassing PromptStyler's accuracy of $73.3$\%. Furthermore, when employing model ensemble, both training time and GPU memory usage remain lower than that of PromptStyler, with consistently superior performance ($74.4\%$ accuracy). Finally, during the testing stage, despite a slight increase in parameter count, our inference speed aligns closely with PromptStyler's and significantly outperforms ZS-CLIP. Note that there is some randomness in the inference speed considering hardware factors, but in general ``DPStyler w/o ME" and ``PromptStyler" have similar inference speed.

\textbf{Style refresh. }We conduct an additional ablation study about style refresh, as presented in TABLE \ref{tab:table_refresh}. Within this study, we compare our standard DPStyler with a variant where we disable the refresh strategy, meaning the style vectors remain unchanged after their initial setup. With the style remover in place and model ensemble applied, we observe the impact of solely disabling the refresh mechanism on the results. It can be seen that without refreshing styles at the beginning of each training epoch, the model's performance declines across various datasets. This indicates that refreshing style vectors can generate a more diverse range of styles.
\begin{table}
\small
\centering
\caption{Comparison with zero-shot CLIP for per-domain top-1 accuracy on NICO++. ZS-CLIP (C) refers to zero-shot CLIP using `[class]' as its text prompt, and ZS-CLIP (PC) indicates zero-shot CLIP using `a photo of a [class]' as its text prompt. \textbf{Bold} indicates the best result.}
\resizebox{\columnwidth}{!}{

\begin{tabular}{l|ccccccc} 
\toprule
\multicolumn{1}{l|}{\multirow{2}{*}{Method}} &   \multicolumn{5}{c}{Accuracy (\%) $\uparrow$}                                                                      \\ 
\cline{2-8}
\multicolumn{1}{c|}{}                                               & \quad Autumn              & Dim              & Grass      & Outdoor        & Rock & Water &  Avg.           \\ 
\hline

\multicolumn{7}{c}{ResNet-50 with pre-trained weights from CLIP}                                                                                                               \\ 
\hline
ZS-CLIP (C) \cite{radford2021learning}                                               &  \quad 79.9           &  74.7          &  84.1           &  78.9   &  81.0   &  68.4           &  77.8            \\
ZS-CLIP (PC) \cite{radford2021learning}                                            &  \quad 81.9           &  77.2           &  86.1           &  80.5   &  83.0 &  71.6          &  80.1           \\  \hline
\rowcolor[rgb]{0.929,0.929,0.929} DPStyler                  &  \quad\textbf{83.6}   &  \textbf{79.4}  &  \textbf{87.1}  &  \textbf{83.7} &  \textbf{84.5}  &  \textbf{76.0} & \textbf{82.4}  \\ 

\hline
\multicolumn{7}{c}{ViT-B / 16 with pre-trained weights from CLIP}                                                                                                              \\ 
\hline
ZS-CLIP (C) \cite{radford2021learning}                                            & \quad 89.7  &  86.6    &   90.9     &  85.7   &    88.7    &   80  &  86.9           \\
ZS-CLIP (PC) \cite{radford2021learning}                                          &    \quad  90.3   & 88.5   &   \textbf{91.6}  &    86.8    &   90.1   &   82.0     &  88.2           \\
\hline
\rowcolor[rgb]{0.929,0.929,0.929} DPStyler                    &  \quad\textbf{91.1}  &  \textbf{88.5}  &  91.4  &  \textbf{87.5} &  \textbf{90.6}   &  \textbf{82.4} &  \textbf{88.6}   \\ 
\hline
\multicolumn{7}{c}{ViT-L / 14 with pre-trained weights from CLIP}                                                                                                              \\ 
\hline
ZS-CLIP (C) \cite{radford2021learning}                                               & \quad 92.0     &89.0 &    92.2    &    85.9   &    90.1   &  83.7   &  88.8           \\
ZS-CLIP (PC) \cite{radford2021learning}                                      & \quad 93.0 &       91.3   &   92.8   &     \textbf{88.2}  &  92.4    & 85.3    &  90.5           \\
\hline
\rowcolor[rgb]{0.929,0.929,0.929} DPStyler                 &  \quad \textbf{93.8}  &  \textbf{91.9}  & \textbf{93.0}  &  88.0      &  \textbf{93.3}       &  \textbf{86.6}         &  \textbf{91.1}             \\
\bottomrule
\end{tabular}
}
\label{tab:table_nico}
\end{table}

\begin{table}[]
\centering
\small
\caption{Ablation experiments on refreshing across PACS, VLCS, OfficeHome (OH), and DomainNet (DN). \textbf{Bold} indicates the best result.}
\resizebox{\columnwidth}{!}{
\begin{tabular}{l|cccc|c} 
\toprule
\multirow{2}{*}{Refresh Strategy} & \multicolumn{5}{c}{Accuracy (\%) $\uparrow$}                                              \\ 
\cline{2-6}
                          & PACS          & VLCS          & OH           & DN            & Avg.           \\ 
\hline
w/o refresh               &    93.3       & 82.8  &    72.1     &  47.1       &    73.8       \\
\rowcolor[rgb]{0.929,0.929,0.929} w refresh   & \textbf{93.6}  &    \textbf{83.5}       & \textbf{72.5}  & \textbf{48.0} &  \textbf{74.4} \\
\bottomrule
\end{tabular}
}
\vspace{-0.2cm}
\label{tab:table_refresh}
\end{table}
\section{conclusion}
In this paper, we propose DPStyler, comprising Style Generation and Style Removal modules. The Style Generation module refreshes all styles at every training epoch, while the Style Removal module captures domain-invariant features. Moreover, we introduce model ensemble to mitigate model sensitivity to input text prompts. Experiments on four benchmark datasets show that DPStyler outperforms recent competitors significantly. Given that our method still has room for improvement on large-scale datasets, we plan to address this issue in future work.

{\small
\bibliographystyle{IEEEtran}
\bibliography{main}
}

\vfill

\end{document}